\documentclass[10pt,journal,compsoc,twoside]{IEEEtran}

\usepackage{amsmath,amsfonts}
\usepackage{graphicx}

\usepackage{booktabs}
\usepackage{tabularx}
\usepackage{multirow}
\usepackage{makecell}
\usepackage{threeparttable}

\usepackage{xcolor}
\usepackage[export]{adjustbox}
\usepackage{fix-cm}
\usepackage{cite}

\usepackage{microtype}

\usepackage[hidelinks]{hyperref}

\newcommand{\papertitle}[0]{[De$\mid$Re]constructing VLMs' Reasoning \\ in Counting}

\newcommand{\clips}[0]{CLIP~}
\newcommand{\clip}[0]{CLIP}
\newcommand{\cocas}[0]{CoCa~}
\newcommand{\coca}[0]{CoCa}
\newcommand{\interns}[0]{InternVL3~}
\newcommand{\intern}[0]{InternVL3}
\newcommand{\llavais}[0]{LLaVA-Interleave~}
\newcommand{\llavai}[0]{LLaVA-Interleave}
\newcommand{\llavaos}[0]{LLaVA-OneVision~}
\newcommand{\llavao}[0]{LLaVA-OneVision}
\newcommand{\paligemmas}[0]{PaliGemma 2~}
\newcommand{\paligemma}[0]{PaliGemma 2}
\newcommand{\qwens}[0]{Qwen2-VL~}
\newcommand{\qwen}[0]{Qwen2-VL}

\begin{document}
\bstctlcite{IEEEexample:BSTcontrol}
\title{\papertitle}
\author{Simone Alghisi, Gabriel Roccabruna\textsuperscript{$\dagger$}\thanks{\textsuperscript{$\dagger$}Work done while at University of Trento, prior to joining Amazon.}, Massimo Rizzoli, Seyed Mahed Mousavi, \\ Giuseppe Riccardi,~\IEEEmembership{Fellow,~IEEE}
\thanks{Simone Alghisi, Gabriel Roccabruna, Massimo Rizzoli, Seyed Mahed Mousavi, and Giuseppe Riccardi are with the SISLab and the Department of Information Engineering and Computer Science, University of Trento, 38123 Povo, Italy. E-mails: s.alghisi@unitn.it, gabriel.roccabruna@unitn.it, massimo.rizzoli@unitn.it, mahed.mousavi@unitn.it, giuseppe.riccardi@unitn.it}
}

\markboth{IEEE Transactions on Pattern Analysis and Machine Intelligence}{Alghisi \MakeLowercase{\textit{et al.}}: [De$\mid$Re]constructing VLMs' Reasoning in Counting}

\IEEEtitleabstractindextext{%
    \begin{abstract}
    Vision-Language Models (VLMs) have recently gained attention due to their competitive performance on multiple downstream tasks, achieved by following user-input instructions. However, VLMs still exhibit several limitations in visual reasoning, such as difficulties in identifying relations (e.g., spatial, temporal, and among objects), understanding temporal sequences (e.g., frames), and counting objects. In this work, we go beyond score-level benchmark evaluations of VLMs by investigating the underlying causes of their failures and proposing a targeted approach to improve their reasoning capabilities. We study the reasoning skills of seven state-of-the-art VLMs in the counting task under controlled experimental conditions. Our experiments show that VLMs are highly sensitive to the number and type of objects, their spatial arrangement, and the co-occurrence of distractors. A layer-wise analysis reveals that errors are due to incorrect mapping of the last-layer representation into the output space. Our targeted training shows that fine-tuning just the output layer improves accuracy by up to 21\%. We corroborate these findings by achieving consistent improvements on real-world datasets. 
    \end{abstract}

    \begin{IEEEkeywords}
    Vision-Language Models, visual reasoning, counting, scene understanding
    \end{IEEEkeywords}
    
    \begin{center}
        \small
        This work has been submitted to the IEEE for possible publication.\\ 
        Copyright may be transferred without notice, after which this version may no longer be accessible.
    \end{center}
}

\maketitle

\section{Introduction}
\noindent \IEEEPARstart{V}{ision} is one of the primary senses through which humans perceive the world. Natural language, in turn, provides a powerful means of describing the relations and interactions of observed events and entities, as well as for conveying abstract concepts, such as ideas, emotions, and instructions. 
Recently, natural language has been employed as an interface for neural models in a multitude of tasks, such as image generation~\cite{Rombach_2022_CVPR, Hu_2024_CVPR}, image editing~\cite{Brooks_2023_CVPR, Huang_2024_CVPR}, and language navigation~\cite{10160969, pmlr-v202-driess23a, Wang_2024_CVPR}. To effectively perform these tasks in dynamic and complex real-world environments, these models are expected to reason on the perceived environments and textual instructions. For these reasons, research on the reasoning skills of Vision-Language Models (VLMs) has been gaining increasing attention from research communities across diverse disciplines.

To evaluate the reasoning capabilities of VLMs, studies have proposed Visual Question Answering (VQA) benchmarks spanning multiple domains (e.g., science, humanities, and business) and targeting specific cognitive skills (e.g., mathematical, spatial, temporal, logical, or commonsense reasoning)~\cite{10.1007/978-3-031-20074-8_9,lu2024mathvista,fu2024blink,al2024unibench,yue-etal-2025-mmmu}. Using VQA benchmarks, previous research has shown that VLMs struggle with basic reasoning tasks, including identifying relations~\cite{liu-etal-2023-visual,fu2024blink}, understanding temporal sequences~\cite{Fu_2025_CVPR,NEURIPS2023_90ce332a}, and counting objects~\cite{campbell2024understanding, rahmanzadehgervi2024vision}. However, most of these studies stop at score-level evaluations, typically reporting only accuracy or F1-score. Only a few have investigated the causes of these failures, relying on datasets containing real-world images~\cite{Tong_2024_CVPR,chen2025why,neo2025towards}. While real-world images represent the desired evaluation setting, they also introduce substantial complexity, often depicting cluttered scenes with multiple objects and occlusions~\cite{rizzoli2025civet, kamath-etal-2023-whats}. Furthermore, such datasets are often unbalanced and usually suffer from inherent biases (e.g., target objects frequently appearing in the center of the image~\cite{kirillov2023segment}), annotation errors, and confounding factors, including ill-posed or ambiguous questions~\cite{DBLP:conf/nips/ChenLDZZCDWQLZ24, yue-etal-2025-mmmu}. This inherent complexity makes it difficult to isolate specific reasoning errors, thereby limiting the interpretability of their actual reasoning performance. Consequently, little is known about \textit{why} VLMs struggle to reason, hindering research on \textit{how} their reasoning capabilities can be enhanced. 

In this work, we move beyond score-level benchmark evaluations by identifying the causes of VLMs' reasoning failures and proposing a targeted approach to improve their reasoning capabilities. We focus our study on VLMs' reasoning skills in counting~\cite{acharya2019tallyqa, Paiss_2023_ICCV, Dai_2024_CVPR}, a representative reasoning task that combines low-level perception with higher-level reasoning. Counting can be decomposed into two stages: \textit{(i)} object identification, which requires visual and semantic understanding of the image to determine the set of target objects, and \textit{(ii)} enumeration, which returns the cardinality of the set by assigning a unique index to each instance~\cite{beckwith1966process, trick1994small, gelman1978child}. This formulation allows us to minimize the prior knowledge required to interpret the image (e.g., by using elementary objects), thereby isolating the reasoning component and enabling a more precise investigation of VLMs' reasoning capabilities. In addition, counting is crucial in many real-world applications, such as precision agriculture (e.g., estimation of crop yields)~\cite{10287390, Matos_2024_CVPR}, traffic monitoring and optimization (e.g., quantification of vehicles or pedestrians)~\cite{Lin_2023_CVPR, 10330623}, and digital pathology (e.g., cell counting)~\cite{song2023artificial, Tyagi_2023_CVPR}.

Using counting as a case study, we frame our work around three research questions:

\textbf{RQ1: Are VLMs capable of reasoning under unbiased input data and evaluation conditions?} Leveraging the CIVET framework~\cite{rizzoli2025civet}, we perform a rigorous and systematic analysis of the counting capabilities of seven state-of-the-art VLMs adopting balanced and unbiased sets of synthetic stimuli. Our results reveal that their counting skills are limited, even when counting up to nine. Further analysis shows that their performance is highly sensitive to external factors, including the class and attributes of the target object, their position in the image, and the presence of distractors. 

\textbf{RQ2: What are the VLM's layers that mainly contribute to their reasoning abilities?} Through a layer-wise analysis, we identify the points of failure within the architecture that affect the model's counting capabilities. Notably, our study identifies the output layer as the main source of errors, due to incorrect mapping of the final layer's representation in the output space. 

\textbf{RQ3: Can we improve the reasoning capabilities of VLMs based on our layer-wise analysis?} We introduce a targeted training approach that enhances VLM's counting capabilities by fine-tuning only the faulty layers. Results show that our computationally efficient approach increases accuracy by up to 21\% on synthetic data.

We summarize our main contributions as follows:
\begin{itemize}
    \item We conduct a comprehensive evaluation of seven state-of-the-art VLMs using balanced and unbiased synthetic stimuli; to understand the cause of the observed errors, we perform a layer-wise analysis to identify VLM's points of failure.
    \item We propose a targeted training approach that only fine-tunes the output layer, yielding up to 21\% higher accuracy.
    \item We validate the effectiveness of the proposed targeted fine-tuning approach on real-world data, demonstrating consistent improvements in performance.
\end{itemize}

\section{Related Works}
\noindent
\textbf{Counting}
Counting refers to the task of determining the number of instances of a target object in an image. Initially, data-driven models were limited to count objects from a specific class~\cite{xie2018microscopy, liang2022end, lin2022boosting} (e.g., ``cells'' or ``people in a crowd''). With the introduction of VLMs, two main strategies have been proposed to generalize across classes: specifying the target object using multiple images (few-shot)~\cite{you2023few, he2024few} or a textual prompt (zero-shot)~\cite{Paiss_2023_ICCV, Jiang_2023_CLIP-Count, qian2025t2icount}. In the latter, VLMs have also been explored for fine-grained counting by incorporating specific attributes in the target description~\cite{Dai_2024_CVPR, triaridis2025improving} (e.g., ``people with sunglasses''). Following this line of research, we specify the class and the attributes of the target object in a zero-shot setting, and evaluate their counting capabilities using image classification~\cite{Paiss_2023_ICCV, al2024unibench} or next-token prediction~\cite{Zhang_2024_CVPR, campbell2024understanding}. 

\textbf{Evaluating VLMs' Reasoning}
VLMs have shown competitive performance on a wide range of tasks, such as classification~\cite{conti2023vocabulary, Saha_2024_CVPR}, segmentation~\cite{kirillov2023segment, 10558790}, and detection~\cite{Li_2022_CVPR, liu2024grounding}. Despite this, existing studies have shown that VLMs struggle even with simple reasoning tasks, such as understanding basic object properties and relations~\cite{rizzoli2025civet, Peng_2024_CVPR}, and counting up to ten~\cite{rahmanzadehgervi2024vision, al2024unibench}. To assess the reasoning capabilities of these models, several VQA benchmarks have been proposed. Some focus on evaluating VLMs across multiple domains~\cite{10.1007/978-3-031-20074-8_9, fu2024blink, yue-etal-2025-mmmu}, while others target specific skills, such as mathematical~\cite{lu2024mathvista}, spatial~\cite{liu-etal-2023-visual} or temporal~\cite{Fu_2025_CVPR} reasoning abilities. Although most studies primarily focus on score-level evaluations, a smaller set of works has investigated the sources of VLM's reasoning limitations.

To understand how these models integrate visual information, a study revealed that masking object-specific tokens could reduce accuracy by up to 70\%~\cite{neo2025towards}. Another study found that attending to these tokens is also crucial for spatial reasoning~\cite{chen2025why}. Analyses of pre-training data revealed that poor spatial reasoning may be attributed to the scarcity and ambiguity of spatial prepositions in large-scale datasets~\cite{kamath-etal-2023-whats}. Further work examined the impact of using CLIP as the vision encoder~\cite{Tong_2024_CVPR}. When evaluated on visually distinct images that are close in CLIP's latent space, many VLMs generated incorrect or hallucinated answers, indicating that their visual shortcomings may stem from CLIP's limitations.

\section{Methodology}
\noindent
Following prior work~\cite{fu2024blink,rahmanzadehgervi2024vision}, we assess the counting capabilities of VLMs by asking questions about an image, and formalize counting as a function\footnote{This definition is partially inspired by the one provided in Referring Expression Counting (REC)~\cite{Dai_2024_CVPR}.}: 
\begin{equation}
    f: (\mathcal{I}, q) \rightarrow \mathbb{N}
\end{equation}

where:
\begin{itemize}
    \item $\mathcal{I} \in \mathbb{R}^{H \times W \times C}$ is an image with height $H$, width $W$, and $C$ channels, depicting some objects.
    \item $q$ is a question in natural language that specifies the \textbf{class} and optionally the \underline{attributes} of the target object (e.g., {\em How many \underline{red} \textbf{circles} are there?}).
    \item $\mathbb{N}$ denotes the set of natural numbers.
\end{itemize}
In other words, $f(\mathcal{I}, q)$ returns the number of objects within image $\mathcal{I}$ that match the class and attributes specified in $q$. 

$\mathcal{I}$ is merely a 3D matrix of raw pixel values, which by itself carries no explicit notion of objects or semantics. To enable reasoning, the image must first be interpreted so that meaningful objects and their attributes can be extracted from this raw signal. We formalize this process as a function $g$:
\begin{equation}
\label{eq:obj_identification}
    g(\mathcal{I}) = \{o_1, o_2, \ldots, o_k\} = \mathcal{O}
\end{equation}
where $g(\mathcal{I})$ returns the multiset\footnote{A multiset allows multiple instances for each of its elements.} $\mathcal{O} = \{o_1, o_2, \ldots, o_k\}$ of objects contained in the image $\mathcal{I}$. Following the definition given in basic formal ontology~\cite{arp2015building}, each \textit{object} $o \in \mathcal{O}$ is an instance of a class and is characterized by a set of attributes. A \textit{class} is understood as a universal, a general type under which multiple objects fall, and typically associated with a noun, such as ``swan'', ``person'', or ``square''. \textit{Attributes}, also referred to as qualities or dependent continuants, are characteristics that inhere in objects, such as mass, color, or shape.

Thus, counting can be detailed as:
\begin{equation}
    \label{eq:count}
    f(\mathcal{I}, q) = \left| \big\{o \in g(\mathcal{I}) \mid match(o, q)\big\} \right|
\end{equation}
where $f(\mathcal{I}, q)$ returns the number of target objects in $\mathcal{I}$, i.e., the objects $o \in g(\mathcal{I})$ that $match$ the object described in $q$. In our definition, two objects match when they belong to the same class and, if specified, share the same attributes.

While the codomain of $f$ is theoretically defined as $[0, +\infty)$, existing evaluation benchmarks constrain valid answers to a finite range to ensure balanced label distributions. For example, CLEVR~\cite{Johnson_2017_CVPR} limits counting to the integer range $[0, 10]$, while datasets such as CountBench~\cite{Paiss_2023_ICCV}, CountBenchQA~\cite{beyer2024paligemmaversatile3bvlm}, and Pixmo-Count~\cite{Deitke_2025_CVPR} restrict the output to the range $[2, 10]$. Although we formalized $f$ as a two-step process where objects $o$ are first extracted from the input image $\mathcal{I}$, most off-the-shelf VLMs address counting as a single end-to-end problem. In practice, given a VLM $\mathcal{M}$, we assess $\mathcal{M}$'s performance by comparing its response $\hat{y} = \mathcal{M}(\mathcal{I}, q)$ with the ground-truth answer $y$. To handle cases where the model outputs may include additional tokens (e.g., ``The answer is\ldots''), we employ a simple regex to match the first valid answer from $\hat{y}$.

\subsection{Controlled and Unbiased Stimuli}
\label{subsec:eval-methodology}
To assess the counting capabilities of VLMs, recent studies have mostly considered datasets containing complex real-world images~\cite{fu2024blink, al2024unibench}. However, the presence of external confounding factors (including occlusions, ill-posed questions, or other forms of ambiguities) and the uneven number of examples per object class may lead to inaccurate estimates of the actual capabilities of these models.

We address these issues by employing the CIVET framework~\cite{rizzoli2025civet}. Using CIVET, we perform a rigorous and systematic evaluation of VLM counting skills by generating balanced and unbiased sets of synthetic stimuli $\mathcal{D}$. Specifically, we design our sets to be balanced in terms of target objects and labels, allowing us to investigate the presence of possible pre-training biases in the models. For each dataset $\mathcal{D}$, we define a stimulus $(\mathcal{I}, q, y) \in \mathcal{D}$ as a triplet, where $\mathcal{I}$ is an image representing a multiset of objects $\mathcal{O}$, $q$ is a question about the content of the image, and $y$ is the corresponding ground-truth answer. With this framework, we can specify the class and attributes of every object $o$, as well as their quantity and position within $\mathcal{I}$. Such fine-grained control allows us to design tailored and deterministic experimental settings, free from occlusions and ill-posed questions. To minimize the variability between images containing $i$ and $i-1$ target objects, we construct the dataset $\mathcal{D}_i$ recursively:
\begin{equation}
\label{eq:dataset}
    \mathcal{D}_i =  \left\{\,\big(adds(\mathcal{I}, o), q, i\big) \mid (\mathcal{I}, q, i{-}1) \in \mathcal{D}_{i-1} \,\right\}
\end{equation}
where:
\begin{itemize}
    \item $\mathcal{I}$ is an image that contains $i{-}1$ objects matching the class and attributes specified in the question $q$.
    \item $o$ is an object that matches the target described in $q$.
    \item $adds(\mathcal{I}, o)$ returns a new image by adding one object $o$ to the multiset $g(\mathcal{I})$ of objects contained in the image $\mathcal{I}$, based on Equation~\ref{eq:obj_identification}.
\end{itemize}
In other words, we generate $\mathcal{D}_i$ by systematically adding one object $o$ to each image $\mathcal{I}$ in $\mathcal{D}_{i-1}$. This ensures that every image in $\mathcal{D}_i$ differs from one in $\mathcal{D}_{i-1}$ by exactly one object, thereby minimizing variations due to object position. 

Sections~\ref{subsec:evaluating} and \ref{sec:datasets} provide additional details on each setting, including the construction of training and test sets and the choices of the key factors under investigation.

\subsection{Layer-Wise Probing}
\label{subsec:methodology-identifying}
Understanding why VLMs struggle to reason is a critical step towards improving their reasoning capabilities. However, identifying the underlying cause of errors is challenging due to the complexity of both the tasks and the models. Existing studies have primarily focused on score-level evaluations~\cite{fu2024blink,al2024unibench,yue-etal-2025-mmmu}, often reporting performance using only automatic metrics, or have relied on complex real-world images~\cite{Tong_2024_CVPR,chen2025why,neo2025towards}, limiting the interpretability of their results.

To address these limitations, we propose a systematic approach that identifies potential sources of errors within the model architecture. Our approach leverages a dataset $\mathcal{D}$ that is balanced and free from confounding factors (Section~\ref{subsec:eval-methodology}). This ensures that errors made by a model $\mathcal{M}$ on a task $t$ can be attributed solely to the limitations of $\mathcal{M}$. Formally, given such an optimal dataset $\mathcal{D}$ used to evaluate $\mathcal{M}$'s performance on a task $t$, we assess whether the representation extracted by a particular layer $l \in \mathcal{M}$ encodes sufficient information to solve $t$. To this end, we recursively construct $\mathcal{H}_i$ as follows:
\begin{equation}
\label{eq:hidden_dataset}
    \mathcal{H}_i = 
    \begin{cases}
      \left\{\,\big(l_0(\mathcal{I}, q), y\big) \mid (\mathcal{I}, q, y) \in \mathcal{D} \,\right\} & \text{if } i = 0 \\\\
      \left\{\,\big(l_i(h_{i-1}), y\big) \mid (h_{i-1}, y) \in \mathcal{H}_{i-1} \,\right\}         & \text{if } i > 0
    \end{cases}
\end{equation}
where:
\begin{itemize}
    \item $(\mathcal{I}, q)$ is an input pair consisting of an image $\mathcal{I}$ and a question $q$ about its content.
    \item $y$ is the ground-truth answer associated with $(\mathcal{I}, q)$
\end{itemize}
In other words, each sample $(h_i, y) \in \mathcal{H}_i$ is a pair, where $h_i$ is the hidden representation extracted by layer $l_i$, and $y$ is the corresponding ground-truth answer.
For $i=0$, $h_0 = l_0(\mathcal{I}, q)$ is the hidden representation produced by the input layer $l_0$ when processing the input $(\mathcal{I}, q)$.
For $i>0$, $h_i = l_i(h_{i-1})$ is the hidden representation produced by layer $l_i$ when processing the representation $h_{i-1}$ of the previous layer.

We can assess whether layer $l_i$ encodes sufficient information to solve the task by measuring the performance of a simple probing model $P_i$ (e.g., a random forest, support vector machine, or linear classifier) trained on $\mathcal{H}_i$. Different from prior methods~\cite{JMLR:v12:montavon11a, DBLP:conf/iclr/AlainB17, pmlr-v235-ghandeharioun24a}, our approach emphasizes training the probing model $P_i$ on balanced data, free from confounding factors. This is a critical design choice that improves the reliability of the evaluation and yields a more accurate estimation of $\mathcal{M}$'s limitations.

\section{Experimental Design}
\label{sec:experimental}
We assess seven state-of-the-art VLMs on the counting task by asking closed-ended questions about an image~\cite{fu2024blink,rahmanzadehgervi2024vision}.

\subsection{Evaluating the Counting Capabilities of VLMs}
\label{subsec:evaluating}
Ideally, counting should not be affected by other irrelevant elements present in the image. In other words, counting ``\textit{red apples}'' should yield the same result when apples are placed alone on a shelf, or when they are mixed with apples of different shapes and colors. Similarly, under experimental conditions, changing the composition of the scene should not affect the model's counting performance.

We test the robustness of VLMs by comparing their performance under different settings. In all settings, we study how varying the class, attributes, and position of the target objects affects their counting capabilities. To ensure that observed failures can be attributed to reasoning limitations, we focus on images containing elementary objects (i.e., 2D shapes) with simple attributes (i.e., color and size), thereby minimizing the prior knowledge to understand the scene. In particular, we consider the combinations of 4 classes (i.e., \textit{square}, \textit{circle}, \textit{triangle}, \textit{star}) and 6 colors (i.e., \textit{red}, \textit{green}, \textit{blue}, \textit{cyan}, \textit{magenta}, \textit{yellow}), for a total of 24 different target objects. 

\begin{figure*}[t]
    \centering
    \includegraphics[width=0.8\linewidth]{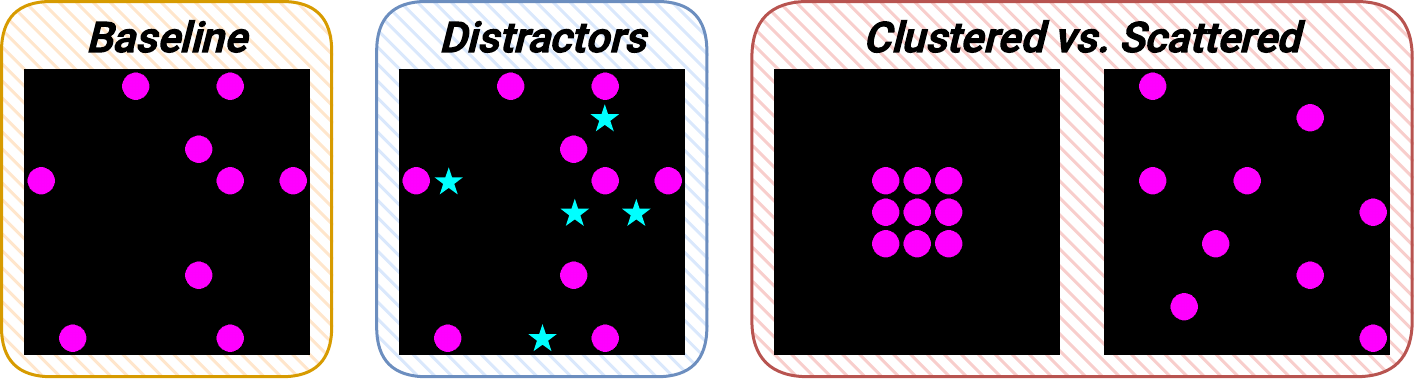}
    \caption{Example images from each experimental setting used to evaluate the counting performance of VLMs under different conditions. In all cases, models are asked to answer the question: \textit{“How many magenta circles are there?”}. The \textbf{Baseline \textit{setting}} includes only target objects. The \textbf{Distractors \textit{setting}} introduces irrelevant objects to assess their impact on model accuracy. The \textbf{Clustered vs. Scattered \textit{setting}} examines whether specific spatial arrangements of objects affect performance. \textit{For clarity, distractors have been depicted in a different color to enhance the contrast with the targets.}}
    \label{fig:datasets}
\end{figure*}

We first analyze the performance of VLMs in a baseline setting, with images containing only target objects. To emulate more realistic scenarios, we then introduce distractors, that is, additional objects that differ in terms of class or attributes from the target object. Finally, we examine the impact of different spatial arrangements by comparing model performance on scenes where target objects are clustered versus scenes in which they are scattered across the image. We provide a visual overview of the samples used in each setting in Figure~\ref{fig:datasets}.

\textbf{Baseline}
We assess the counting performance of VLMs in a baseline setting with images containing only target objects. By eliminating non-target objects from the scene, we ensure that variations in counting performance can be attributed solely to different combinations of class, attributes, and positions of the target objects.

\textbf{Distractors}
To investigate the capabilities of VLMs towards more realistic settings, we measure the influence of distractors on model accuracy. Specifically, we assess the effect of the number of distractors, their class and attributes, and their position in the image. Since exhaustively testing all combinations of targets and distractors is computationally infeasible, we focus our analysis on images where the targets are ``\textit{large magenta circles}'', which have been shown to yield the highest accuracy, as illustrated in Table~\ref{tab:target_object_performance}.

For this setting, we generate our samples by adding the distractors on top of the baseline images (as illustrated in Figure~\ref{fig:datasets}). This ensures that any difference in performance between the two settings is due to the presence of distractors. We first test each model using a distractor that differs both in terms of class and attributes from the target. Based on the baseline results in Table~\ref{tab:target_object_performance}, we select ``\textit{small red star}'' as a distractor, since it is the object with the highest accuracy that satisfies our requirements. Intuitively, as a distractor becomes more similar to the target object, the counting task should become increasingly difficult. To evaluate this hypothesis, we gradually increase the similarity between the distractor (i.e., ``\textit{small red stars}'') and the target (i.e., ``\textit{large magenta circles}'') by varying individual properties. Specifically, we consider the following distractors: ``\textit{large red stars}'' (sharing size), ``\textit{large red circles}'' (sharing size and shape), and ``\textit{large magenta stars}'' (sharing size and color).

\textbf{Clustered vs. Scattered}
When objects are placed too close together, their boundaries may overlap, causing multiple objects to be perceived as one. Since VLMs process visual inputs by dividing images into patches, clustered objects may be grouped within a single patch. When counting, this could cause the model to underestimate the number of target objects within a visual scene.
To investigate this phenomenon, we consider two contrasting spatial configurations illustrated in Figure~\ref{fig:datasets}: the \textit{clustered} setting, where objects are localized in a region, and the \textit{scattered} setting, where objects are distributed more uniformly across the image.

\subsection{Identifying Failure Points in VLMs}
\label{subsec:identifying}
To understand why VLMs struggle to count, we conduct a layer-wise analysis to identify specific points of failure within a VLM $\mathcal{M}$, as formalized in Section~\ref{subsec:methodology-identifying}. Our investigation focuses on encoder-decoder architectures (such as \paligemma~\cite{steiner2024paligemma2familyversatile}, \qwen~\cite{Qwen2VL}, and \intern~\cite{zhu2025internvl3exploringadvancedtraining}), as they have generally achieved the highest performance in our evaluation (as shown in Table~\ref{tab:baseline}). To ensure that counting errors can be attributed only to $\mathcal{M}$'s limitations, we use the dataset $\mathcal{D}$ designed for our baseline setting. In the context of encoder-decoder VLMs, the hidden representation $h_i$ (Section~\ref{subsec:methodology-identifying}) is a sequence obtained by concatenating the visual and textual information at the $i$-th layer, i.e., $h_i = V_i \bigoplus T_i$. Here, $V_i$ and $T_i$ are the representations of the visual and textual tokens at layer $l_i$.
At the input layer, $V$ corresponds to the visual tokens produced by the vision encoder after being projected into the embedding space of the decoder, while $T$ corresponds to the embedding of the textual tokens.

Due to its high dimensionality, training $P_i$ directly on $h_i$ is computationally demanding and may lead to suboptimal or misleading results (since high-dimensional data may often be trivially separable). We thus consider different strategies to aggregate the information contained in $h_i$:
\begin{itemize}
    \item $H_{\text{last}}$: we select the hidden representation for the final token of $h_i$, which is expected to summarize the preceding context effectively (due to the attention mechanism).
    \item $\overline{H}$: to investigate whether a uniform and global representation of the sequence might be useful for the task, we consider the arithmetic mean of all hidden representations in $h_i$.
    \item $V_{\text{last}}$: since $V_i$ should capture the necessary information about the objects in the image, we extract the representation of the last visual token.
    \item $\overline{V}$: analogous to $\overline{H}$, we consider the mean of $V_i$ as an alternative to $V_{\text{last}}$.
\end{itemize}

To understand whether the vision encoder may be the source of the errors, we also train $P_0$ on the average encoder representation $Enc$.
Furthermore, following prior work~\cite{geva-etal-2023-dissecting, mousavi2025llms}, we examine the contribution of intermediate layers by projecting the final token $H_{\text{last}}$ through the output layer $Out$ and measuring its accuracy.

In our experiments, we use a linear Support Vector Machine (SVM) as the probing model $P_i$, implemented via scikit-learn's \texttt{SVC} class~\cite{scikit-learn} with a linear kernel and regularization parameter set to 1. This choice ensures the approach is computationally efficient, interpretable, and minimally sensitive to hyperparameter tuning. We also employ 3-fold cross-validation to ensure robust and unbiased performance estimates.

\subsection{Improving VLM's counting capabilities}
\label{subsec:improving}
We propose a targeted training approach to enhance the reasoning capabilities of VLMs. To leverage the outcomes of our layer-wise analysis, we fine-tune only the layers $l_i$ that, based on the probing model $P_i$, are negatively affecting the model. We then quantify the effectiveness of our targeted training approach by comparing the fine-tuned model with its pre-trained counterpart on a separate test set.

We investigate whether VLMs can learn to count in a simplified environment, isolated from unwanted confounding factors. Similar to the baseline setting described in Section~\ref{subsec:evaluating}, we construct a balanced training dataset with images containing only elementary target objects. To understand whether our findings are not dependent on our synthetic data, we also fine-tune and evaluate each model using real-world counting datasets. This step allows us to assess whether the improvements observed in the synthetic setting transfer to more realistic scenarios, characterized by higher complexity (e.g., occlusions and distractors).
Further details on the training and test datasets are provided in Section~\ref{sec:datasets}.

\section{Datasets}
\label{sec:datasets}
We construct controlled synthetic datasets for model training and evaluation, detailing the visual and textual stimuli (i.e., images and questions). In addition, we adapt Pixmo-Count \cite{Deitke_2025_CVPR} into a balanced benchmark to assess VLM counting performance on real-world images. We provide an overview of these datasets in Table~\ref{sec:datasets}.

\begin{table}[t]
    \caption{Synthetic and real-world datasets used for training and evaluating VLMs' counting capabilities. The table reports the number of samples and target objects for each dataset split.}
    \label{tab:datasets}
    \centering
    \begin{threeparttable}
        \begin{tabular}{lrrrrrr}
            \toprule
            \multirow{2}{*}{\textbf{Dataset}} 
            & \multicolumn{3}{c}{\textbf{\# Samples}} 
            & \multicolumn{3}{c}{\textbf{\# Target Objects}} \\
            \cmidrule(l{2pt}r{3pt}){2-4} \cmidrule(l{3pt}r{2pt}){5-7}
            & \makecell{Train} & \makecell{Valid} & \makecell{Test} 
            & \makecell{Train} & \makecell{Valid} & \makecell{Test} \\
            \midrule
            \textbf{Synthetic}                  & & & & & & \\
            \quad \textit{Training}             & $4{,}860$ & $2{,}430$ & --     & $10$   & $10$   & -- \\
            \quad \textit{Baseline}             & --     & --     & $17{,}496$ & --   & --   & $24$ \\
            \quad \textit{Distractors}          & --     & --     & $26{,}244$ & --   & --   & $1$ \\
            \quad \textit{Clustered}            & --     & --     & $9{,}408$  & --   & --   & $24$ \\
            \quad \textit{Scattered}            & --     & --     & $9{,}408$  & --   & --   & $24$ \\\\
            \textbf{Real-World}                       & & & & & & \\
            \quad \textit{BPC}                  & $3{,}000$ & $480$   & $480$   & $76$  & $51$  & $60$ \\ 
            \bottomrule
        \end{tabular}
    \end{threeparttable}
\end{table}

\subsection{Synthetic Stimuli}
\label{subsec:data}
In our experiments, we represent each image $\mathcal{I}$ as a 9x9 grid, where objects with different classes and attributes can be placed inside a cell. To prevent irrelevant elements from being mistaken as target objects, we consider images with a black background\footnote{Since the color black carries zero information, it is effective to reduce the relevance of the background in the scene understanding.}. Furthermore, we place each object in a distinct cell to eliminate ambiguities due to occlusions. While the resulting synthetic images may be out of distribution with respect to the pre-training data, this setup minimizes the influence of potential non-controllable confounding factors, allowing us to focus solely on the counting capabilities of these models. Investigating the impact of different backgrounds is left as future work.

Each closed-ended question $q$ prompts the model to count the target objects in an image $\mathcal{I}$, specifying the object class and attributes. The set of possible answer options is then appended to the question in the form: \textit{“Choose from [$<$options$>$]”}, where \textit{$<$options$>$} is a comma-separated list of all possible answers. To mitigate order bias, we shuffle these options so that each possible order appears uniformly across the textual inputs. We also reduce the likelihood of open-ended or verbose responses by instructing the model to \textit{``Answer with as few words as possible''}, based on previous findings~\cite{rizzoli2025civet}. The Supplementary Material provides additional details about the data used in the proposed work.

We design multiple synthetic datasets that enable controlled evaluation of VLMs under different settings, as discussed in Section~\ref{subsec:evaluating}. For each setting, we add objects and distractors incrementally, as described in Section~\ref{subsec:eval-methodology}. In addition, we consider images containing at most nine target objects. This choice is supported by the fact that different VLMs have different vocabularies, meaning that the same corpus could be split into tokens and sub-tokens in different ways. For instance, ``42'' could be either tokenized into a single token or as ``4'' followed by ``2''. This discrepancy leads to an unfair comparison: for a vocabulary of size $v$, tokenizing a number as a single token gives a random-guess probability of $1/v$, while splitting it into two tokens reduces the probability to $1/v^2$.

\textbf{Baseline} We construct a dataset $\mathcal{D} = \{\mathcal{D}_1\, \cup \ldots \cup \,\mathcal{D}_9\}$ with images containing from one to nine objects (i.e., in the range $[1, 9]$). To ensure a balanced distribution across object counts, each subset $\mathcal{D}_i$ contains exactly 81 samples, matching the number of examples for $i=1$, which has the lowest cardinality (see Section~\ref{subsec:eval-methodology}).

We study how different target objects affect the performance of VLMs by creating one version of $\mathcal{D}$ for each of our 24 elementary objects. Specifically, for each triplet $(\mathcal{I}, q, a) \in \mathcal{D}$, we modify the class and attributes of the target object in the image $\mathcal{I}$ and adjust the question $q$ based on the elementary object. As motivated in Section~\ref{subsec:eval-methodology}, we minimize variations due to object placement by keeping the target position fixed across all versions.

\textbf{Distractors} 
 As detailed in Section~\ref{subsec:evaluating}, we construct four versions of the dataset, one for each type of distractor. To improve interpretability, we generate three variants of each image $\mathcal{I} \in \mathcal{D}$ by varying the position of the distractors. Additionally, we assess whether increasing the number of distractors can affect performance by considering images with $1$, $5$, and $9$ distractors. 

\textbf{Clustered vs. Scattered} We construct a dataset for the clustered setting, by placing objects to minimize their average distance from the cluster centroid, and one for the scattered setting, by enforcing that each object is placed at least three cells away from every other object. Different from the previous settings, we consider images containing between two and nine objects (i.e., $[2, 9]$), since a single object could ambiguously represent both configurations. Furthermore, we balance each dataset by considering $49$ samples for each $\mathcal{D}_i$ ($i \in [2, 9]$), as this corresponds to the number of unique images in $\mathcal{D}_9$ when clustering objects (based on our definition). 

\textbf{Training} We construct a synthetic dataset to enhance the counting abilities of VLMs, following the same procedure as the baseline setting. To prevent overfitting, we change the target objects' positions and their class-attribute combinations. In total, we consider ten target objects: four \textit{white} shapes (i.e., \textit{square}, \textit{circle}, \textit{triangle}, \textit{star}) and six \textit{plus}-shaped objects in different colors (i.e., \textit{red}, \textit{green}, \textit{blue}, \textit{cyan}, \textit{magenta}, \textit{yellow}). After fine-tuning on our synthetic training set, we evaluate the models on the baseline setting to quantify their improvement.

\subsection{Real-World data}
\label{subsec:real-data}
To corroborate the validity of our findings beyond the synthetic setting, we consider training and evaluating our models using real-world images. Since older datasets may have become part of the (pre-)training data of some VLMs, we focus on the recently released Pixmo-Count~\cite{Deitke_2025_CVPR}. Pixmo-Count comprises 38k images, automatically annotated using standard object detectors. Specifically, its images contain between 0 and 255 target objects (with an average of $3.66 \pm 3.33$ objects per image), and its questions cover 365 unique object classes (with $65.92\%$ focusing on the class "people"). Similar to the approach proposed in CountBenchQA~\cite{Paiss_2023_ICCV, beyer2024paligemmaversatile3bvlm}, the authors manually verified 120 samples for each object count in the range [2, 10], creating a balanced validation and test set (with 540 images each). Despite these efforts, the training set remains heavily imbalanced in terms of object count, as the number "3" appears more frequently than "8", and object class, since the class "people" occurs in two-thirds of the questions, which may lead a model to overfit on the most frequent pattern. Moreover, since only 76 of the 365 object classes appear in the validation and test sets, training on the full dataset introduces numerous irrelevant classes. To address these issues, we construct the Balanced Pixmo-Count (BPC) dataset by uniformly sampling images across all 76 target object categories. Furthermore, we ensure uniform label distribution by sampling 300 images for each object count in the range [0, 9]\footnote{We include images with 0 and 1 objects, as some object classes appear exclusively in this range.}. Consistent with Section~\ref{subsec:data}, we restrict the Pixmo-Count validation and test sets to images containing at most 9 target objects (i.e., using the range [2, 9] instead of the original [2, 10]). The resulting BPC dataset consists of 3,000 training samples, 480 validation samples, and 480 test samples.

\section{Models}
\noindent
We select seven models spanning diverse architectures and training strategies. We categorize these models into two main categories, \textit{dual-encoder} and \textit{encoder-decoder} architectures. \textit{Dual-encoder} models consist of separate vision and text encoders, typically trained to maximize the similarity between image-caption pairs. Within this category, we consider \clips ViT-L/14-336px~\cite{pmlr-v139-radford21a} as a baseline, and \cocas~\cite{ilharco_gabriel_2021_5143773, yucoca} as an improved version of CLIP that complements contrastive pre-training with a captioning objective. Since dual-encoders lack generative capabilities, we reformulate our VQA task as a retrieval problem, where each answer option is treated as a candidate class. Following standard zero-shot image classification using \clip~\cite{pmlr-v139-radford21a}, we encode both the input image and each answer option and select the option with the highest similarity (i.e., dot product) with the image.

\textit{Encoder-decoder} models combine a vision encoder with a text decoder. The vision encoder first extracts a visual representation of the input image. This representation is then projected into the decoder's embedding space. The final result is concatenated with the text embedding and jointly processed by the text decoder. These models are typically pre-trained on image captioning tasks and subsequently fine-tuned on downstream tasks such as VQA, object detection, and segmentation. For this category, we evaluate \llavais 7B\cite{li-2025-llava}, \llavaos 7B\cite{lillava}, \qwen-7B-Instruct\cite{Qwen2VL}, \paligemmas 10B\cite{steiner2024paligemma2familyversatile}, and \interns 8B~\cite{zhu2025internvl3exploringadvancedtraining}. Although all encoder-decoder models share a common architecture, they differ in how visual information is projected into the decoder's embedding space. \llavai, \llavao, \paligemma, and \interns employ a projection layer to align the visual encoder with the text decoder. In the case of LLaVA models, only the projection layer is trained, whereas in \paligemmas and \interns the entire architecture is fine-tuned. In \qwens, the vision encoder and text decoder are directly combined, as they share the same output and input dimensions, respectively. These components are jointly pre-trained, allowing them to learn a shared representation. Regarding image processing, most models rely on vision encoders that accept inputs of a fixed resolution. For instance, both LLaVA models require images of exactly $384\times384$ pixels, while \paligemmas and \interns use $448\times448$. To process high-resolution images, \paligemmas downsamples them to the required size, potentially losing fine-grained details, while \llavai, \llavao, and \interns subdivide them into smaller patches, which may split crucial information across non-sequential tokens. By contrast, \qwens is capable of natively handling variable image resolutions, eliminating the need for explicit resizing or cropping. Additional details about the checkpoint used for each model can be found in the Supplementary Material.

\section{Results}
\noindent 
We evaluate seven state-of-the-art VLMs on counting tasks across multiple settings (RQ1), identify their limitations via layer-wise inspection (RQ2), and show the effectiveness of our targeted training approach (RQ3). We primarily quantify performance using accuracy, as all datasets are balanced in terms of labels. Additional results based on Mean Absolute Error (MAE) and Root Mean Squared Error (RMSE) are presented in the Supplementary Material.

\subsection{RQ1: Are VLMs capable of reasoning under unbiased input data and evaluation conditions?}
\label{subsec:rq1}

\textbf{Baseline}
We report the performance of the models under the baseline setting in Table~\ref{tab:baseline}, computing accuracy across the 24 elementary target objects\footnote{Additional preliminary results on these elementary objects can be found in the Supplementary Material.} defined in Section~\ref{subsec:evaluating}. Since a target object is an instance of a class (e.g., star or circle) and is characterized by a set of attributes (i.e., its color), we analyze the effect of varying the class of the target object by marginalizing over all attributes. For each class $c_j$, we compute the accuracy $acc_j$ using only samples containing target objects of class $c_j$, thus marginalizing over attribute variations. The variability across classes is then quantified by reporting the standard deviation over $acc_1, acc_2, \ldots, acc_n$. An analogous procedure is applied to evaluate the influence of the target object's attributes on model performance.

\begin{table}[ht]
    \caption{Accuracy ($\%$) under the baseline setting, with standard deviations (STD) quantifying the impact of varying the class (e.g., star or circle) and the attributes (e.g., color) of the target object.}
    \label{tab:baseline}
    \centering
    \begin{threeparttable}
        \begin{tabular}{lccc}
            \toprule
            \textbf{Model} & \textbf{Accuracy} & \textbf{Class (STD)} & \textbf{Attributes (STD)} \\
            \midrule
            \vspace{0.5em}
            \textit{Random Baseline} & $11.11$ &              &            \\
            \textbf{Dual-Encoders}          & & & \\
            \quad \textit{\clip}           & $20.52$ &  $\pm 9.35$  & $\pm 2.16$ \\
            \vspace{0.5em}
            \quad \textit{\coca}           & $48.16$ &  $\pm 7.78$  & $\pm 3.19$ \\
            \textbf{Encoder-Decoders}          & & & \\
            \quad \textit{\intern}         & $52.32$ &  $\pm 11.51$ & $\pm 2.23$ \\
            \quad \textit{\llavai}         & $45.66$ &  $\pm 4.79$  & $\pm 1.08$ \\
            \quad \textit{\llavao}         & $67.46$ &  $\pm 6.57$  & $\pm 1.68$ \\
            \quad \textit{\paligemma}      & $74.86$ &  $\pm 1.54$  & $\pm 1.46$ \\
            \quad \textit{\qwen}           & $64.72$ &  $\pm 6.87$  & $\pm 1.63$ \\
            \bottomrule
        \end{tabular}
        {\footnotesize \textit{Overall accuracy is computed across all 24 elementary target objects. For each model, the STD for class (or attributes) is computed by marginalizing over attributes (or classes).} \par}
    \end{threeparttable}
\end{table}

Among the models, \paligemmas achieves the highest accuracy ($74.86\%$), while \clips yields the lowest ($20.52\%$). Although encoder-decoder models generally achieve better results, \llavais performs worse than \coca, suggesting that counting capabilities are not dependent on the category of the VLM. Instead, the substantial gaps in performance observed among encoder-decoder models, including the $29.2\%$ difference between \paligemmas and \llavai, suggest that the choice of vision encoder, text decoder, or pre-training data is more important.

Analysis of the standard deviations reveals that changing the class of the target object has a greater impact on model performance than varying its attributes. \interns exhibits the highest class-level variability, implying that its counting capability depends heavily on the target's class. In terms of attributes, \cocas shows the highest sensitivity, followed by \interns and \llavai. Among all models, \paligemmas is the least affected by class variations, indicating more robust counting capabilities. One potential explanation is the inclusion of TallyQA~\cite{acharya2019tallyqa}, a well-established counting dataset, in its pre-training mixture, as reported in the model paper~\cite{beyer2024paligemmaversatile3bvlm, steiner2024paligemma2familyversatile}. However, verifying this hypothesis is difficult given the limited transparency about the pre-training data mixtures of other VLMs. The influence of the class on the model performance is also supported by Table~\ref{tab:target_object_performance}, which reports the counting accuracy for each target object, averaged across all models. Each entry corresponds to a specific target object, defined by the combination of class (circles, squares, stars, triangles) and attribute (red, green, blue, cyan, magenta, yellow). Consistent with the trends observed in Table~\ref{tab:baseline}, performance varies substantially when changing the class of the object, with the gap between ``circles'' and ``squares'' exceeding $8.8\%$. Among all target objects, \textit{``magenta circles''} achieve the highest counting accuracy, whereas \textit{``cyan squares''} yield the lowest.

\begin{table}[ht]
    \caption{Accuracy ($\%$) of each target object under the baseline setting. A target object is defined as the combination of class (circles, squares, stars, triangles) and attribute (red, green, blue, cyan, magenta, yellow).}
    \label{tab:target_object_performance}
    \centering
    \begin{threeparttable}
        \begin{tabular}{lcccc}
            \toprule
            & \textbf{Circles} & \textbf{Squares} & \textbf{Stars} & \textbf{Triangles} \\
            \midrule
            \textbf{Red}    & $57.67$           & $48.21$ & \underline{$57.46$} & $54.65$ \\
            \textbf{Green}  & $55.32$           & $46.52$ & $53.67$ & $51.71$ \\
            \textbf{Blue}   & $57.85$           & $48.64$ & $55.22$ & $54.77$ \\
            \textbf{Cyan}   & $58.34$           & $45.15$ & $55.03$ & $53.07$ \\
            \textbf{Magenta}& $\mathbf{59.22}$  & $47.62$ & $55.44$ & $54.32$ \\
            \textbf{Yellow} & $57.44$           & $48.78$ & $52.52$ & $52.67$ \\
            \bottomrule
        \end{tabular}
        {\footnotesize \textit{Results are averaged across all models. The best-performing class-attribute combination is shown in \textbf{bold}, and the second-best with a different class and attribute is \underline{underlined}.} \par} 
    \end{threeparttable}
\end{table}

\begin{figure}[ht]
    \centering
    \includegraphics[width=\linewidth]{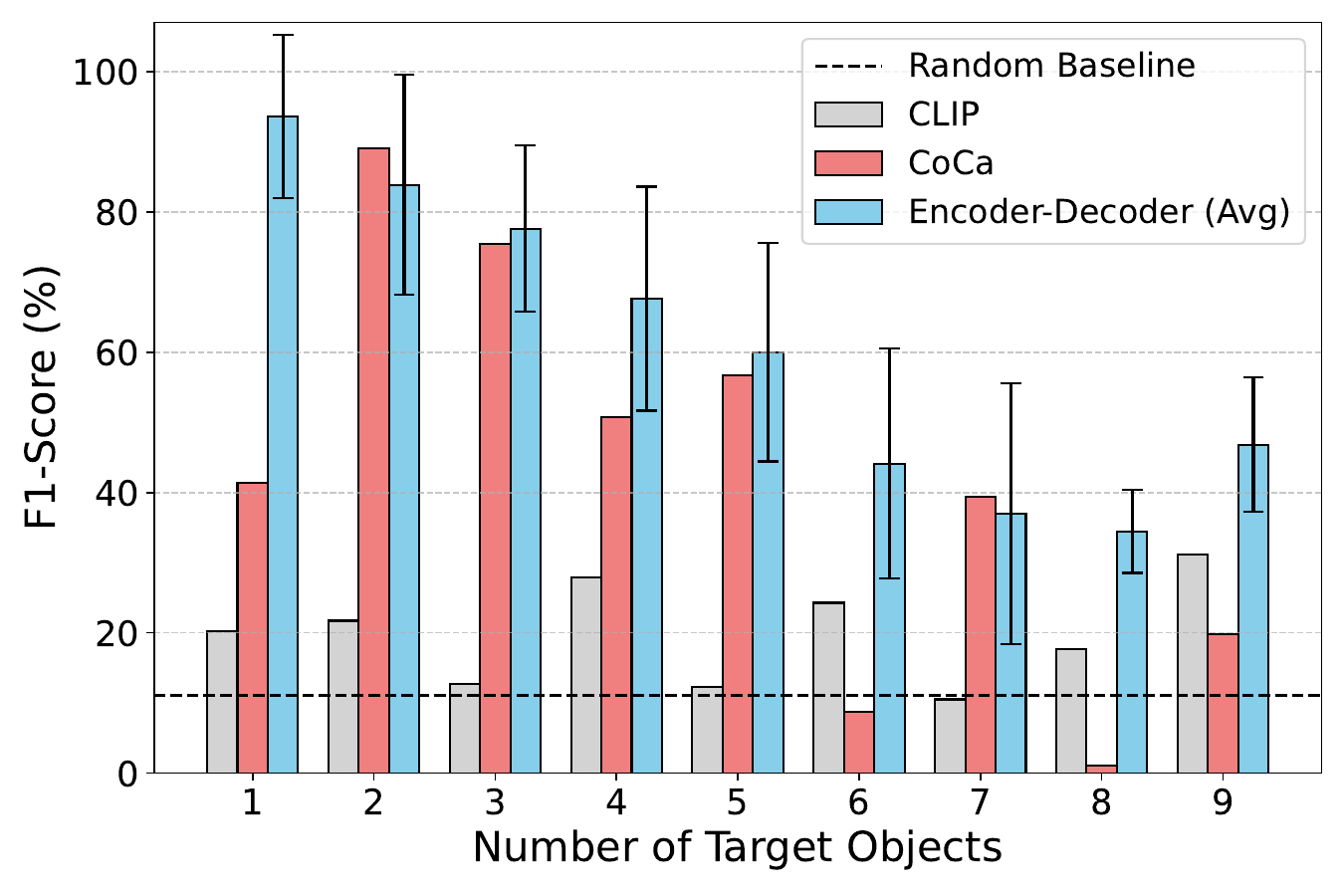}
    \caption{F1-scores under the baseline setting for \clip, \coca, and the Encoder-Decoder models when increasing the number of target objects. Results for Encoder-Decoder models are averaged across \llavai, \llavao, \paligemma, and \qwen, with the corresponding standard deviation reported using error bars.}
    \label{fig:accuracy_per_number_of_target_objects}
\end{figure}

\begin{table*}[ht]
    \caption{Accuracy ($\%$) of each model under the distractors setting when varying the type and number (\#) of the distractors. Except for \intern, performance drops as the target and the distractors become more similar and as the number of distractors increases.}
    \label{tab:distractors}
    \centering
    \begin{threeparttable}
        \setlength{\tabcolsep}{7.5pt}
        \begin{tabular}{lcccccccc}
            \toprule
            \multirow{2}{*}{\textbf{Model}} & \multicolumn{4}{c}{\textbf{Distractor Types}} & \multicolumn{3}{c}{\textbf{\# Distractors}} \\
            \cmidrule(l{4pt}r{4pt}){2-5} \cmidrule(l{4pt}r{4pt}){6-8}
             & \makecell{\texttt{SRS}} & \makecell{\texttt{LRS}} & \makecell{\texttt{LRC}} & \makecell{\texttt{LMS}}  & \makecell{\texttt{1}} & \makecell{\texttt{5}} & \makecell{\texttt{9}} \\
            \midrule
            \textit{\clip}      & 17.48 {\scriptsize \textcolor{red}{(↓ 0.49)}} & 16.17 {\scriptsize \textcolor{red}{(↓ 1.80)}} & 14.11 {\scriptsize \textcolor{red}{(↓ 3.86)}} & 15.15 {\scriptsize \textcolor{red}{(↓ 2.82)}} & 15.99 {\scriptsize \textcolor{red}{(↓ 1.98)}} & 15.32 {\scriptsize \textcolor{red}{(↓ 2.65)}} & 15.88 {\scriptsize \textcolor{red}{(↓ 2.09)}} \\
            \textit{\coca}      & 23.87 {\scriptsize \textcolor{red}{(↓ 34.84)}} & 14.97 {\scriptsize \textcolor{red}{(↓ 43.74)}} & 9.77 {\scriptsize \textcolor{red}{(↓ 48.94)}} & 14.46 {\scriptsize \textcolor{red}{(↓ 44.25)}} & 26.98 {\scriptsize \textcolor{red}{(↓ 31.73)}} & 10.79 {\scriptsize \textcolor{red}{(↓ 47.92)}} & 9.53 {\scriptsize \textcolor{red}{(↓ 49.18)}} \\
            \textit{\intern}    & 68.31 {\scriptsize \textcolor[HTML]{228B22}{(↑ 7.54)}} & 67.49 {\scriptsize \textcolor[HTML]{228B22}{(↑ 6.72)}} & 58.57 {\scriptsize \textcolor{red}{(↓ 2.19)}} & 63.60 {\scriptsize \textcolor[HTML]{228B22}{(↑ 2.83)}} & 61.67 {\scriptsize \textcolor[HTML]{228B22}{(↑ 0.90)}} & 65.07 {\scriptsize \textcolor[HTML]{228B22}{(↑ 4.30)}} & 66.75 {\scriptsize \textcolor[HTML]{228B22}{(↑ 5.98)}} \\
            \textit{\llavai}    & 38.61 {\scriptsize \textcolor{red}{(↓ 15.03)}} & 37.65 {\scriptsize \textcolor{red}{(↓ 15.99)}} & 30.42 {\scriptsize \textcolor{red}{(↓ 23.21)}} & 30.09 {\scriptsize \textcolor{red}{(↓ 23.55)}} & 36.98 {\scriptsize \textcolor{red}{(↓ 16.66)}} & 32.91 {\scriptsize \textcolor{red}{(↓ 20.72)}} & 32.68 {\scriptsize \textcolor{red}{(↓ 20.95)}} \\
            \textit{\llavao}    & 57.95 {\scriptsize \textcolor{red}{(↓ 15.71)}} & 54.17 {\scriptsize \textcolor{red}{(↓ 19.49)}} & 37.89 {\scriptsize \textcolor{red}{(↓ 35.77)}} & 50.08 {\scriptsize \textcolor{red}{(↓ 23.58)}} & 60.73 {\scriptsize \textcolor{red}{(↓ 12.93)}} & 48.50 {\scriptsize \textcolor{red}{(↓ 25.16)}} & 40.83 {\scriptsize \textcolor{red}{(↓ 32.83)}} \\
            \textit{\paligemma} & 66.68 {\scriptsize \textcolor{red}{(↓ 9.86)}} & 64.15 {\scriptsize \textcolor{red}{(↓ 12.39)}} & 56.59 {\scriptsize \textcolor{red}{(↓ 19.95)}} & 60.39 {\scriptsize \textcolor{red}{(↓ 16.16)}} & 70.03 {\scriptsize \textcolor{red}{(↓ 6.52)}} & 60.19 {\scriptsize \textcolor{red}{(↓ 16.36)}} & 55.65 {\scriptsize \textcolor{red}{(↓ 20.90)}} \\
            \textit{\qwen}      & 43.73 {\scriptsize \textcolor{red}{(↓ 29.52)}} & 48.91 {\scriptsize \textcolor{red}{(↓ 24.34)}} & 45.51 {\scriptsize \textcolor{red}{(↓ 27.74)}} & 46.01 {\scriptsize \textcolor{red}{(↓ 27.24)}} & 45.47 {\scriptsize \textcolor{red}{(↓ 27.78)}} & 45.20 {\scriptsize \textcolor{red}{(↓ 28.05)}} & 47.45 {\scriptsize \textcolor{red}{(↓ 25.80)}} \\
            \midrule               
            \textbf{Average}    & 45.23 {\scriptsize \textcolor{red}{(↓ 13.99)}} & 43.36 {\scriptsize \textcolor{red}{(↓ 15.86)}} & 36.12 {\scriptsize \textcolor{red}{(↓ 23.10)}} & 39.97 {\scriptsize \textcolor{red}{(↓ 19.25)}} & 45.41 {\scriptsize \textcolor{red}{(↓ 13.81)}} & 39.71 {\scriptsize \textcolor{red}{(↓ 19.51)}} & 38.40 {\scriptsize \textcolor{red}{(↓ 20.82)}} \\                                               
            \bottomrule
        \end{tabular}
        {\footnotesize \textit{In all experiments, models are asked to count \textit{``How many magenta circles are there?''}. Arrows (\textcolor[HTML]{228B22}{↑} / \textcolor{red}{↓}) indicate absolute increase or decrease in accuracy with respect to the baseline setting. Similar to the baseline, accuracy for each distractor type (or count) is computed by marginalizing over all distractor counts (or types). Notation: \texttt{SRS} = Small Red Star, \texttt{LRS} = Large Red Star, \texttt{LRC} = Large Red Circle, and \texttt{LMS} = Large Magenta Star.} \par} 
    \end{threeparttable}
\end{table*}

\begin{figure}[ht]
    \centering
    \includegraphics[width=\linewidth]{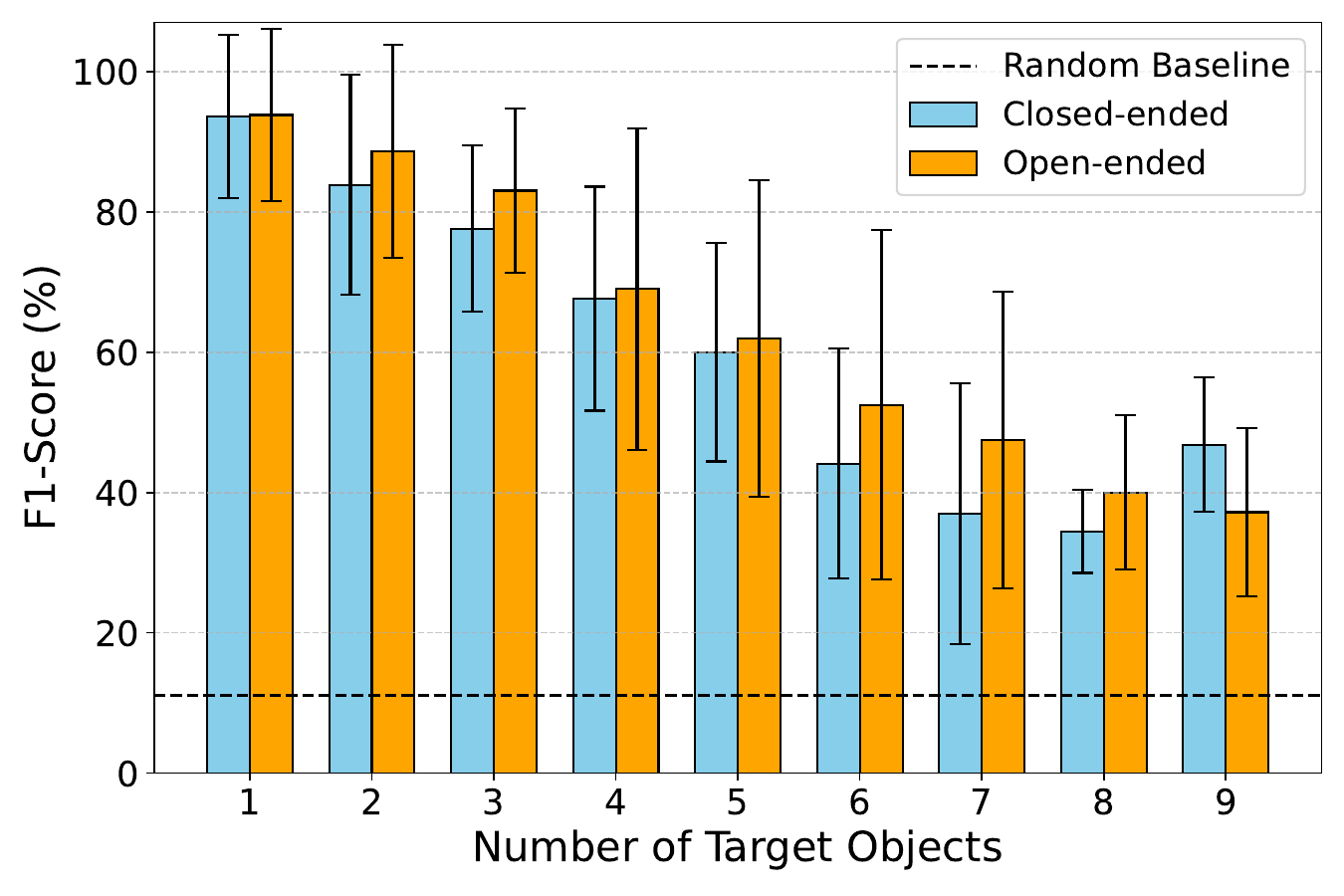}
    \caption{F1-scores under the baseline setting for different numbers of target objects when querying encoder-decoder models with open- or closed-ended questions. Results are averaged across encoder-decoder models with the corresponding standard deviation indicated using error bars.}
    \label{fig:open_vs_close_per_number_of_target_objects}
\end{figure}

To gain further insight into the counting capabilities of VLMs, we analyze performance as a function of the number of target objects, as depicted in Figure~\ref{fig:accuracy_per_number_of_target_objects}. Due to the high number of models, we average the results for encoder-decoder models and report their standard deviation using error bars. Our analysis shows that, although \clips achieved the lowest accuracy in Table~\ref{tab:baseline}, its performance is significantly better than \cocas when processing images containing `6', `8', and `9' objects. For \cocas, while performance generally declines when increasing the number of objects, the model exhibits a lower F1-score even when processing images containing only a single object. This behavior could be due to data imbalance, since some digits appear more frequently than others in the caption used to pre-train VLMs~\cite{acharya2019tallyqa, al2024unibench}. However, this seems not to be the reason, as encoder-decoder models do not suffer from this issue. Indeed, encoder-decoder models exhibit a different pattern: the F1-score for images with `9' objects is higher than `6', `7', and `8'.  Since we hypothesize that closed-ended questions may be the reason, we test the model using open-ended questions and report the results in Figure~\ref{fig:open_vs_close_per_number_of_target_objects}. The findings confirm our hypothesis, showing that performance consistently decreases as the number of objects increases. Moreover, we also observe that open-ended questions increase the overall accuracy of \paligemmas and \interns by $4\%$, suggesting that providing the list of possible answers may instead reduce the model's accuracy.

\textbf{Distractors} 
The results for the distractor setting, described in Section~\ref{subsec:evaluating}, are shown in Table~\ref{tab:distractors}. Consistent with the baseline setting, we report the accuracy for each distractor type by marginalizing over images containing 1, 3, and 5 distractors. The same procedure is applied when evaluating accuracy across different numbers of distractors. For each experiment, we also report the accuracy difference with the baseline performance, computed using images containing only ``\textit{large magenta circles}''.

On average, distractors reduce accuracy, and the effect is stronger when the similarity between the distractor and the target object increases, validating the hypothesis discussed in Section~\ref{subsec:evaluating}. This trend is clear when comparing ``\textit{small red stars}'' (\texttt{SRS}), which share no similarities with the target object, with ``\textit{large red circles}'' (\texttt{LRC}), which have the same size and class. Although ``\textit{large magenta stars}'' (\texttt{LMS}) also share two elements (size and color), their impact is smaller than \texttt{LRC}. Among the models, \cocas ($-48.94\%$) and \llavaos ($-35.77\%$) are the most negatively influenced by the distractor type, while \interns is the only model that benefits from the introduction of distractors. A similar trend is observed as the number of distractors increases, with \cocas ($-49.18\%$) and \llavaos ($-32.83\%$) exhibiting the largest performance drops. Our results reveal a clear negative correlation between the number of distractors and the counting accuracy: the higher the number of distractors, the lower the accuracy. \interns remains the only exception, as it exhibits the opposite trend.

\textbf{Clustered vs. Scattered}
We evaluate the model on images in which objects are either scattered or clustered and report the results in Table~\ref{tab:clustered_scattered}. Overall, models exhibit higher accuracy for the clustered setting, contradicting the hypothesis presented in Section~\ref{subsec:evaluating}. These findings suggest that models may struggle in the scattered setting as they have to combine information from multiple parts of the image. However, this is not the case when considering \clips and \paligemma, which exhibit higher accuracy when the objects are scattered. Finally, the difference in accuracy between the clustered and scattered settings depends on the model, ranging from $1.37\%$ for \llavaos to $17.74\%$ for \paligemma.

\begin{table}[ht]
    \caption{Accuracy ($\%$) of each model under the clustered and scattered settings. Results show that most models struggle to aggregate information when objects are scattered throughout the image.}
    \label{tab:clustered_scattered}
    \centering
    \begin{threeparttable}
        \begin{tabular}{lcc}
            \toprule
            \textbf{Model} & \textbf{Clustered} & \textbf{Scattered} \\
            \midrule
            \textit{\clip}      & $25.10$ & $\mathbf{27.46}$ \\
            \textit{\coca}      & $\mathbf{66.45}$ & $61.08$ \\
            \textit{\intern}    & $\mathbf{66.85}$ & $55.12$ \\
            \textit{\llavai}    & $\mathbf{57.30}$ & $44.08$ \\
            \textit{\llavao}    & $\mathbf{73.00}$ & $72.37$ \\
            \textit{\paligemma} & $59.82$ & $\mathbf{77.56}$ \\
            \textit{\qwen}      & $\mathbf{65.32}$ & $62.56$ \\
            \bottomrule
        \end{tabular}
        {\footnotesize \textit{Accuracy is computed across all 24 elementary target objects} \par}
    \end{threeparttable}
\end{table}

\subsection{RQ2: What are the VLM’s layers that mainly contribute to their reasoning abilities?}
To better understand the source of the limitations observed in Section~\ref{subsec:rq1}, we perform a layer-wise analysis by training linear probing models on the baseline dataset using different intermediate representations, as discussed in Section~\ref{subsec:identifying}.

\begin{figure}[t]
    \centering
    \includegraphics[width=0.8\linewidth]{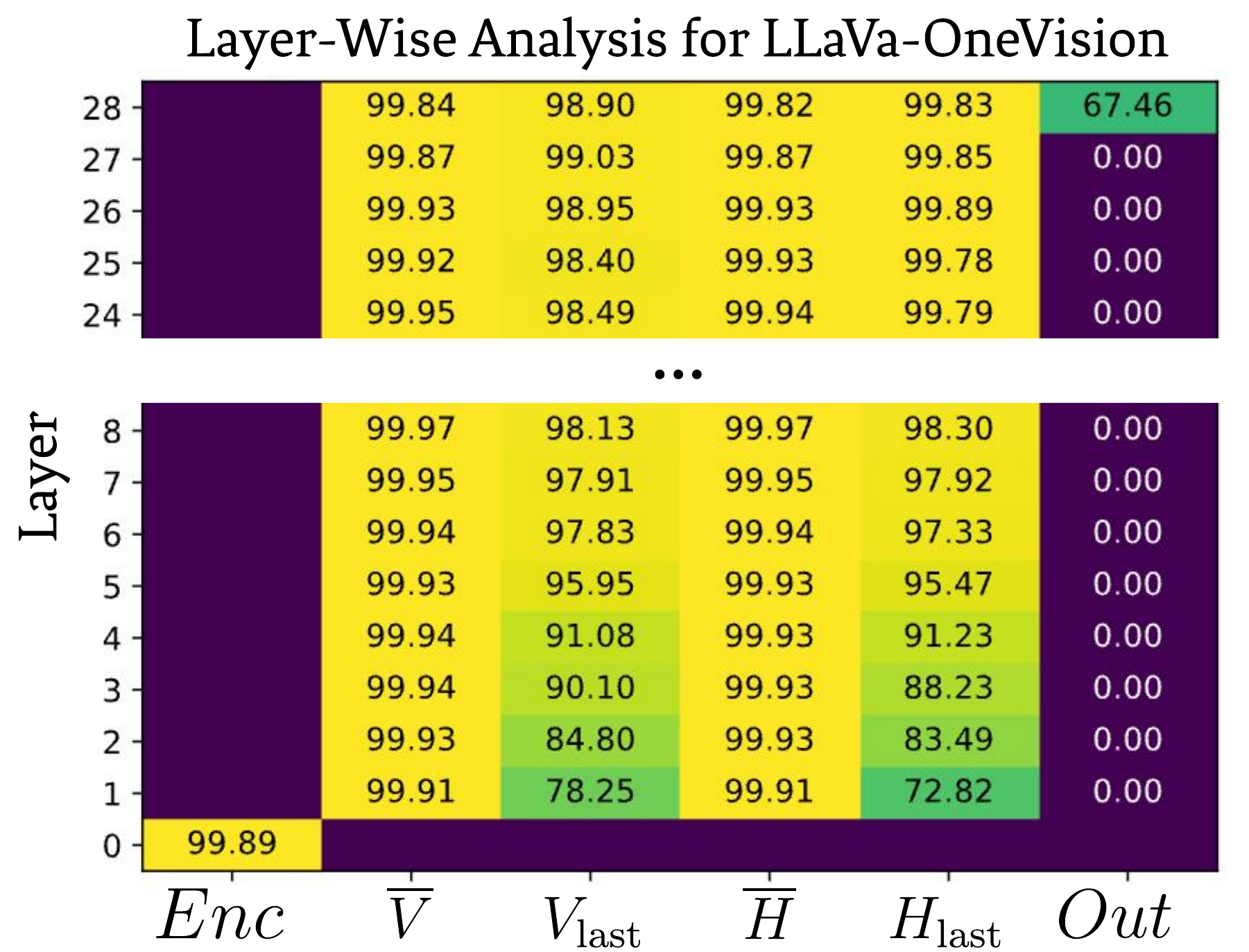}
    \caption{Layer-wise analysis of \llavaos under the baseline setting. For each layer $l_i$, we report the counting accuracy of the probing model $P_i$ when trained on different intermediate representations, as discussed in Section~\ref{subsec:identifying}. Results show that the output layer is the main source of errors, since training $P_i$ on the last token representation $H_{\text{last}}$ yields near-perfect accuracy.
    }
    \label{fig:layer-wise}
\end{figure}

Due to resource constraints, we focus our analysis on encoder-decoder VLMs, as they generally outperform the other models, as shown in Table~\ref{tab:baseline}. We report the results of our layer-wise analysis for \llavaos in Figure~\ref{fig:layer-wise}, focusing on the most meaningful layers. The full set of figures, encompassing all layers and models, is included in the Supplementary Material. We compute the accuracy for $Out$ using greedy decoding, and average the results for $Enc$, $\overline{V}$, $V_{\text{last}}$, $\overline{H}$, and $H_{\text{last}}$ across 3-fold cross-validation. For these representations, the standard deviation remains below 1\%, confirming the stability of the results. In general, our analysis shows that the encoder representation $Enc$ contains sufficient information to directly solve our task, suggesting that the primary problem lies in the decoder. Starting from the lower layers of the decoder, we observe that the models trained on the average representations $\overline{V}$ achieve a higher accuracy than their last-token counterparts $V_{\text{last}}$. However, as information flows to the upper layers, the gap decreases, indicating that the decoder progressively aggregates meaningful information into $V_{\text{last}}$. A similar trend can be observed when comparing $\overline{H}$ with $H_{\text{last}}$. Additionally, models trained on the last layer representation of $H_{\text{last}}$ achieve near-perfect accuracy. When the same representation $H_{\text{last}}$ is passed to the output layer $Out$, \llavaos obtains $30\%$ less accuracy. As our layer-wise analysis reveals that all encoder-decoder models exhibit similar behaviors, we conclude that the output layer is the main source of errors in these VLMs.

\subsection{RQ3: Can we improve the reasoning capabilities of VLMs based on our layer-wise analysis?}
Leveraging the results of our layer-wise analysis, we fine-tune only the output layer while freezing the rest of the parameters. Our targeted training strategy is computationally efficient, enabling us to fine-tune all models on a single GeForce RTX 2080 Ti GPU with 11GiB of memory. Further implementation details are provided in the Supplementary Material.

\begin{figure*}[ht]
    \centering
    \includegraphics[width=0.9\linewidth]{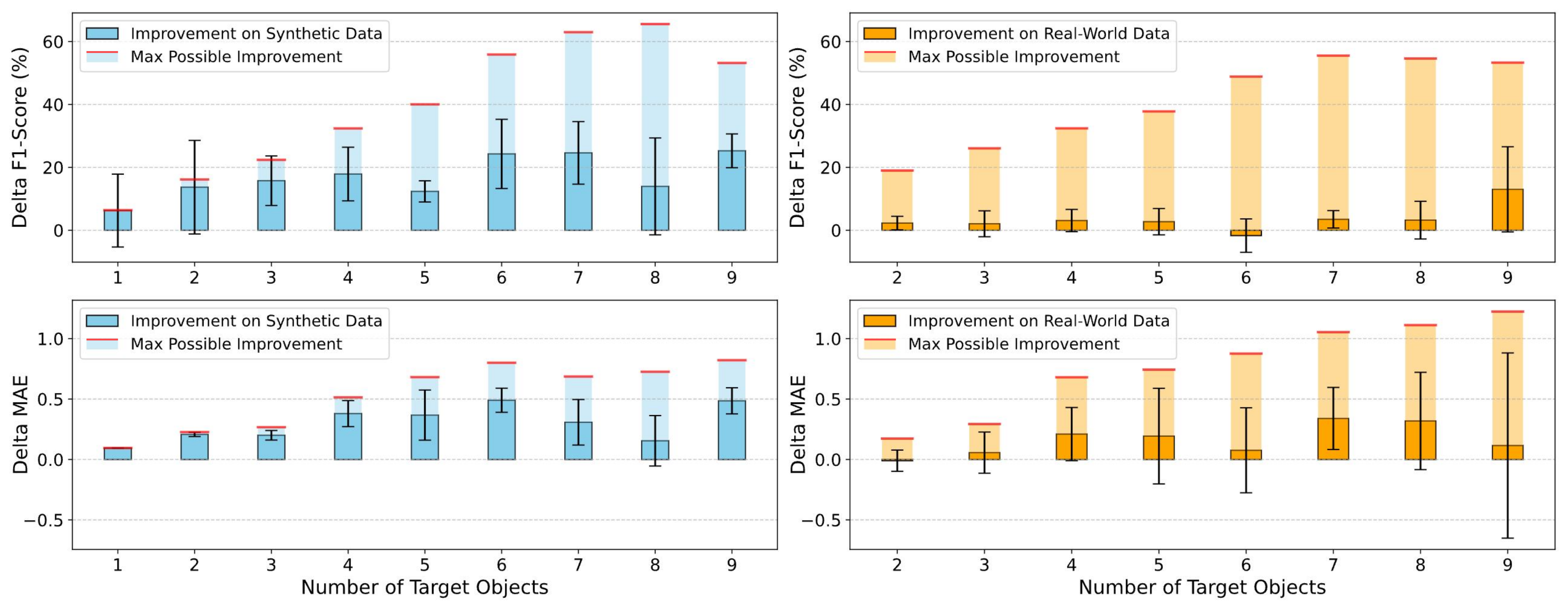}
    \caption{Change in F1-Score (\%) and Mean Absolute Error (MAE) after fine-tuning on synthetic (left) and real-world (right) data as a function of the number of target objects. The number of target objects ranges from 1 to 9 in the synthetic setting and from 2 to 9 in the real-world setting. Bars indicate average improvement across encoder-decoder models, with error bars showing standard deviation. The shaded regions represent the maximum possible improvement relative to the performance of the pre-trained models.}
    \label{fig:fine-tune}
\end{figure*}

\begin{table}[ht]
    \caption{Accuracy ($\%$) and Mean Absolute Error (MAE) on synthetic and real-world images after targeted fine-tuning. Results show consistent improvement across both settings.}
    \label{tab:fine-tuning-acc-mae}
    \centering
    \begin{threeparttable}
        \begin{adjustbox}{width=\columnwidth}
        \begin{tabular}{lcccc}
            \toprule
            \multirow{2}{*}{\textbf{Model}} & \multicolumn{2}{c}{\textbf{Synthetic}} & \multicolumn{2}{c}{\textbf{Real-World}} \\
            \cmidrule(lr){2-3} \cmidrule(lr){4-5}
             & \textbf{Accuracy ↑} & \textbf{MAE ↓} & \textbf{Accuracy ↑} & \textbf{MAE ↓} \\
            \midrule
            \textit{\intern}    & 73.52 {\scriptsize \textcolor[HTML]{228B22}{(21.19)}} & 0.28 {\scriptsize \textcolor[HTML]{228B22}{(0.34)}} & 70.49 {\scriptsize \textcolor[HTML]{228B22}{(3.40)}} & 0.34 {\scriptsize \textcolor[HTML]{228B22}{(0.18)}} \\
            \textit{\llavai}    & 66.63 {\scriptsize \textcolor[HTML]{228B22}{(20.97)}} & 0.39 {\scriptsize \textcolor[HTML]{228B22}{(0.62)}} & 50.96 {\scriptsize \textcolor[HTML]{228B22}{(4.68)}} & 0.88 {\scriptsize \textcolor[HTML]{228B22}{(0.15)}} \\
            \textit{\llavao}    & 81.10 {\scriptsize \textcolor[HTML]{228B22}{(13.64)}} & 0.19 {\scriptsize \textcolor[HTML]{228B22}{(0.18)}} & 49.68 {\scriptsize \textcolor[HTML]{228B22}{(3.61)}} & 1.08 {\scriptsize \textcolor[HTML]{228B22}{(0.19)}} \\
            \textit{\paligemma} & 89.28 {\scriptsize \textcolor[HTML]{228B22}{(14.42)}} & 0.11 {\scriptsize \textcolor[HTML]{228B22}{(0.16)}} & 74.31 {\scriptsize \textcolor[HTML]{228B22}{(7.64)}} & 0.29 {\scriptsize \textcolor[HTML]{228B22}{(0.20)}} \\
            \textit{\qwen}      & 80.09 {\scriptsize \textcolor[HTML]{228B22}{(15.37)}} & 0.21 {\scriptsize \textcolor[HTML]{228B22}{(0.19)}} & 65.61 {\scriptsize \textcolor[HTML]{228B22}{(2.98)}} & 0.43 {\scriptsize \textcolor[HTML]{228B22}{(0.09)}} \\
            \midrule
            \textbf{Average}    & 78.13 {\scriptsize \textcolor[HTML]{228B22}{(17.13)}} & 0.24 {\scriptsize \textcolor[HTML]{228B22}{(0.30)}} & 62.21 {\scriptsize \textcolor[HTML]{228B22}{(4.46)}} & 0.61 {\scriptsize \textcolor[HTML]{228B22}{(0.16)}} \\
            \bottomrule
            \vspace{-2mm}
        \vspace{0.1em}
        \end{tabular}
        \end{adjustbox}
    \end{threeparttable}
    {\footnotesize
    \textit{Values in parentheses indicate absolute improvement with respect to the corresponding pre-trained models.}
    \par}
\end{table}

Table~\ref{tab:fine-tuning-acc-mae} presents the results of our targeted fine-tuning approach, averaged across all target objects. Following prior work on object counting~\cite{Dai_2024_CVPR, Jiang_2023_CLIP-Count}, we also report the MAE, which quantifies the average absolute deviation between predicted and ground-truth counts. We fine-tune each model on both synthetic and real-world data. In the synthetic setting, we fine-tune the models on the training dataset and quantify their improvement over the baseline setting. For the real-world setting, we adopt the newly proposed BPC dataset for both training and evaluation.

On \textbf{synthetic data}, \interns exhibits the largest improvement in accuracy ($+21.19\%$), closely followed by \llavais ($+20.98\%$). In terms of MAE, \llavais yields the largest reduction ($-0.62$), indicating more accurate and consistent count predictions after fine-tuning. \paligemmas maintains the highest overall performance, reaching nearly $90\%$ accuracy with an MAE of $0.11$. As shown in the top-left part of Figure~\ref{fig:fine-tune}, accuracy gains are observed across all object counts, with the largest improvements for images containing `6', `7', and `9' target objects. Nevertheless, since `9' had one of the largest accuracy gain, the characteristic trend highlighted in Figures~\ref{fig:accuracy_per_number_of_target_objects} and \ref{fig:open_vs_close_per_number_of_target_objects} remains visible. This indicates that these biases are not localized in the output layer, but are instead more deeply rooted within the model. A similar pattern can also be observed in the bottom-left part of Figure~\ref{fig:fine-tune}, where the improvement for `9' is higher than `7' and `8'.

On \textbf{real-world data}, \paligemmas achieves the highest performance, reaching $74.31\%$ accuracy with an MAE of $0.20$, followed by \interns at $70.49\%$ accuracy and an MAE of $0.34$. The relatively strong baseline of \interns on real-world images suggests that the model may have been exposed to visually crowded scenes during the pre-training. This could also explain why adding more distractors increases the model's accuracy, as observed in Table~\ref{tab:distractors}. As shown in the top-right part of Figure~\ref{fig:fine-tune}, models fine-tuned on real-world images benefit from our approach, despite achieving more modest accuracy gains. The only exception is for images containing `6' target objects, which exhibit a slight decrease in accuracy after fine-tuning. Notably, while `9' displays the largest accuracy improvement, `7' and `8' have more stable gains in terms of MAE, as shown in the bottom-right part of Figure~\ref{fig:fine-tune}.

Overall, our targeted training approach improves the counting performance of all models on both synthetic ($+17.13\%$) and real-world data ($+4.46\%$). The performance gap between the two settings can likely be attributed to the greater scene complexity inherent to real-world images and to class imbalance, as some target objects are still underrepresented in the BPC dataset (Section~\ref{subsec:real-data}). Some pre-training biases persist even after fine-tuning the output layer, highlighting the need for novel pre-training approaches that disentangle the understanding of basic visual concepts from the acquisition of specific reasoning skills.

\section{Conclusion}
\noindent
In this work, we conducted a rigorous and systematic analysis of the reasoning capabilities of seven state-of-the-art VLMs on the counting task. Unlike previous studies, we move beyond score-level benchmark evaluations by identifying the reasons behind their limitations and proposing concrete strategies to improve their capabilities. Using counting as a case study and the CIVET framework to ensure balanced and unbiased stimuli, we perform an in-depth analysis that provides interpretable and accurate results. Our findings reveal that VLMs exhibit limited counting capabilities, with performance heavily influenced by the class and attributes of the target object, their position, and the presence of distractors. Through our layer-wise analysis, we identify the output layer as the primary source of errors. Building on this insight, we introduce a targeted training strategy, which is computationally efficient and improves counting accuracy by up to $21\%$ on synthetic data. To corroborate the validity and the effectiveness of our results, we apply the same strategy to real-world images and observe consistent improvements. As a future direction, we plan to address the persistence of some pre-training biases by investigating novel training paradigms, designed to equip VLMs with built-in reasoning capabilities. Additionally, we plan to extend our evaluation to consider other types of backgrounds and more realistic scenarios.

\bibliographystyle{IEEEtran}
\bibliography{custom}

\begin{IEEEbiography}[{\includegraphics
[width=1in,height=1.25in,clip,
keepaspectratio]{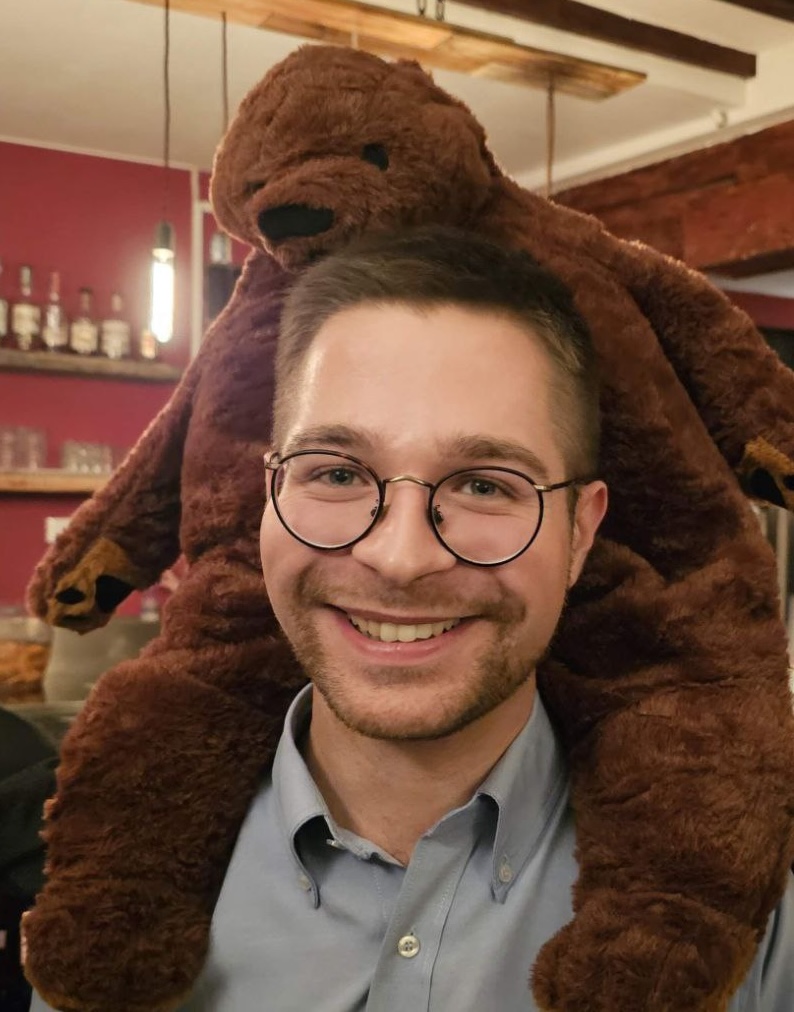}}]
{Simone Alghisi}
is a Ph.D. Candidate at the University of Trento, Italy, in the Signals and Interactive Systems Lab (SISL). He received the M.Sc. degree in Computer Science from the University of Trento in 2023. His research interests include scene understanding, multimodal systems, vision- and large language models. Previously, he was a research intern at Mila - Quebec AI Institute.
\end{IEEEbiography}

\vspace{-1em}

\begin{IEEEbiography}[{\includegraphics
[width=1in,height=1.25in,clip,
keepaspectratio]{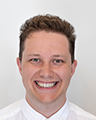}}]
{Gabriel Roccabruna} earned his Ph.D. in Computer Science from the University of Trento, Italy, in 2025. His doctoral research focused on natural language understanding in narratives, with an emphasis on modeling sentiment, emotions, and events. He is currently a full-time Applied Scientist at Amazon in Madrid, Spain. Before this role, he worked as a Research Assistant at the University of Trento for one year and completed an internship at Amazon in Seattle, USA, in 2024.
\end{IEEEbiography}

\vspace{-1em}

\begin{IEEEbiography}[{\includegraphics
[width=1in,height=1.25in,clip,
keepaspectratio]{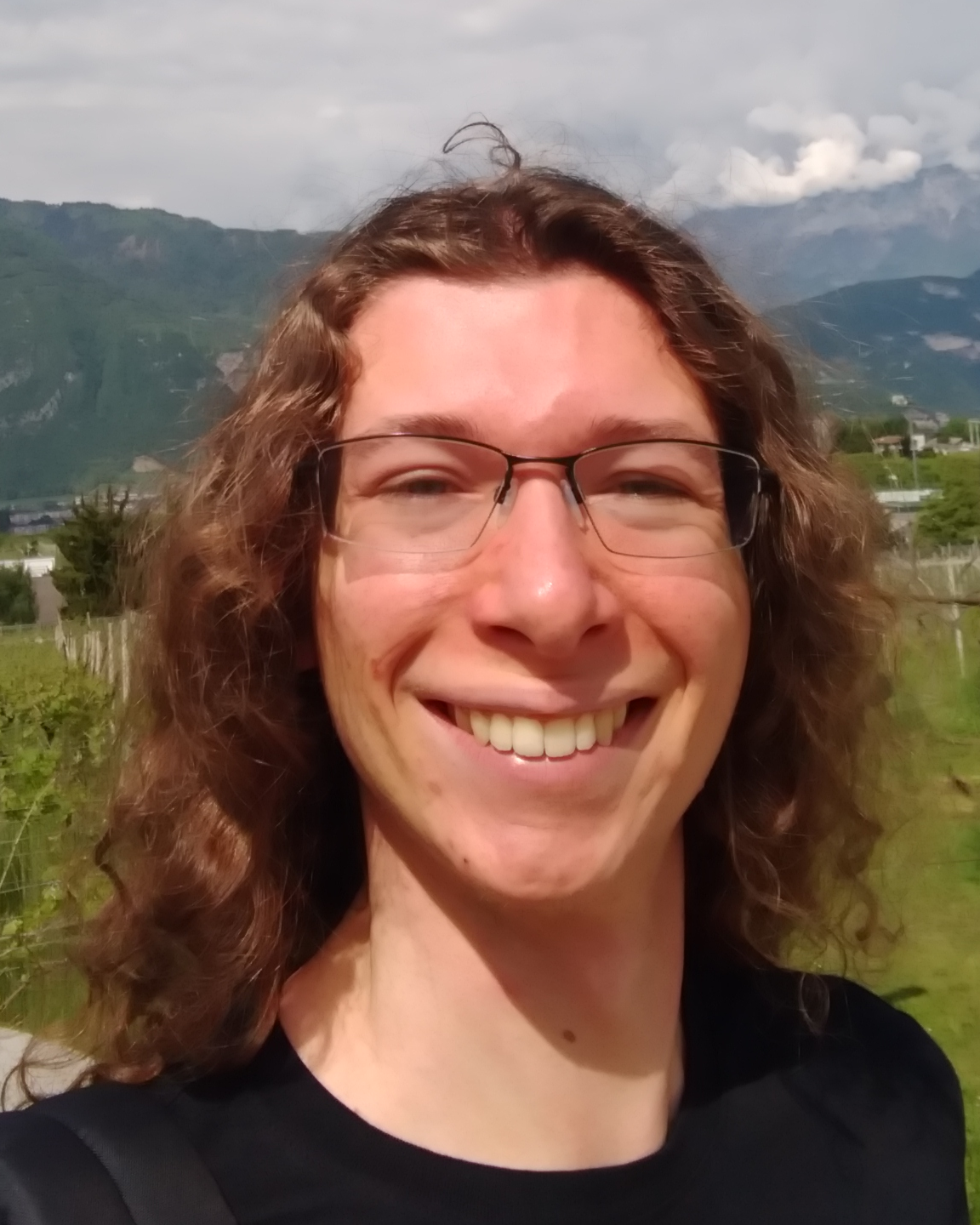}}]
{Massimo Rizzoli}
Massimo Rizzoli is a Ph.D. candidate in Computer Science and Information Engineering at the University of Trento, Italy.
He holds an M.Sc. in AI Systems and completed a research internship at Mila - Quebec AI Institute.
His research interests include natural language processing and multimodal scene understanding, and he has published work on these topics in leading conferences, including EMNLP.
\end{IEEEbiography}

\vspace{-1em}

\begin{IEEEbiography}[{\includegraphics
[width=1in,height=1.25in,clip,
keepaspectratio]{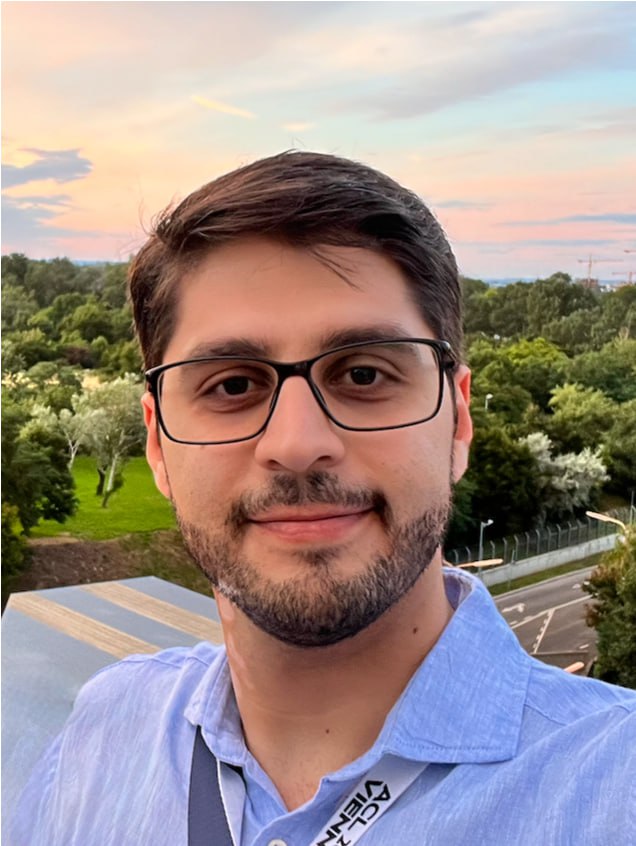}}]
{(S.) Mahed Mousavi} received the Ph.D. degree (cum laude) in Information and Communication Technology from the University of Trento, Italy, in 2023. He is currently an Assistant Professor in the Department of Information Engineering and Computer Science at the University of Trento. He has held visiting researcher positions at Mila-Quebec AI Institute and the University of Helsinki. Dr. Mousavi’s research interests include conversational AI, large language models, and human-centered NLP. His work investigates the robustness, ethics, and social implications of dialogue systems.
\end{IEEEbiography}

\vspace{-1em}

\begin{IEEEbiography}[{\includegraphics
[width=1in,height=1.25in,clip,
keepaspectratio]{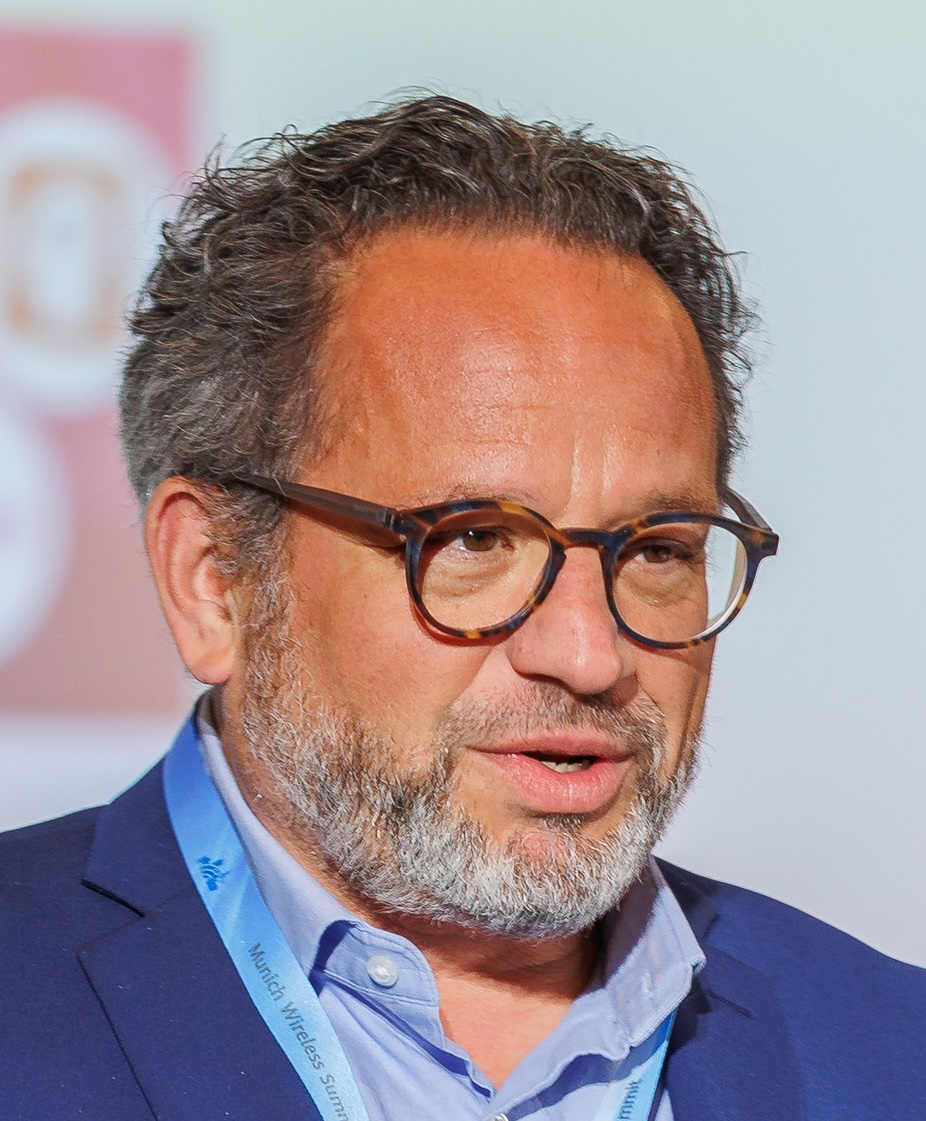}}]
{Giuseppe Riccardi}
(M'96–SM'04–F'10) is the founder and director of the Signals and Interactive Systems Lab at the University of Trento, Italy. He received the MSEE and Ph.D. in Electrical Engineering from the University of Padua (Italy). From 1993 to 2005, he was at AT\&T Bell Laboratories (USA) and then AT\&T Labs-Research (USA) in the Speech and Language Processing Research Division. In 2005, he joined the Department of Information Engineering and Computer Science at the University of Trento (Italy). Prof. Riccardi’s research focuses on conversational artificial intelligence, computational models for natural language processing, human-machine dialogue, and affective computing. In the late 1990s, he and his colleagues at AT\&T designed, trained, and deployed ``How May I Help You ?'', the first large-scale natural language human-machine dialogue system. More recently, he has deployed the first RCT trial of conversational AI for application in the health domain. His research aims to create human-machine interaction and conversational systems that can benefit individuals, their well-being, and their health. He is elected Fellow of ISCA (2017) and AAIA (2023).
\end{IEEEbiography}

\clearpage
\renewcommand{\thefigure}{S\arabic{figure}}
\renewcommand{\thetable}{S\arabic{table}}
\setcounter{figure}{0}
\setcounter{table}{0}

\section*{Supplementary Material}
The supplementary material includes additional experimental results, figures, and implementation details that complement the main manuscript.

\subsection*{Code}
The source code and the complete set of experimental results are publicly available at \url{https://github.com/sislab-unitn/DeRe-VLM}.

\subsection*{Models}
For our experiments, we used the following model checkpoints available from HuggingFace:
\begin{itemize}
    \item \href{https://huggingface.co/openai/clip-vit-large-patch14}{\clips ViT-L/14-336px}
    \item \href{https://huggingface.co/OpenGVLab/InternVL3-8B-hf}{\intern-8B}
    \item \href{https://huggingface.co/llava-hf/llava-interleave-qwen-7b-dpo-hf}{\llavais 7B DPO}
    \item \href{https://huggingface.co/llava-hf/llava-onevision-qwen2-7b-ov-hf}{\llavaos 7B}
    \item \href{https://huggingface.co/google/paligemma2-10b-mix-448}{\paligemmas 10B 448px}
    \item \href{https://huggingface.co/Qwen/Qwen2-VL-7B-Instruct}{\qwen-7B-Instruct}
\end{itemize}
The only exception is CoCa, for which we employed the CoCa ViT-L/14 model (using the \texttt{mscoco\_finetuned\_laion2B-s13B-b90k} checkpoint) provided by the Open-CLIP repository\footnote{\url{https://github.com/mlfoundations/open_clip}}.

\subsection*{Preliminary Results on Scene Understanding}
Since the data used to pre-train VLMs is often undisclosed, it is unclear whether our elementary target objects are known to the models or fall outside their training distribution. To avoid evaluating VLMs on unseen or rare objects, we first conduct a preliminary experiment to assess the scene understanding capabilities of the seven state-of-the-art models considered in this work. We evaluate their performance on 24 elementary target objects, focusing on recognition of object classes (i.e., shapes) and attributes (i.e., colors). Details about the data and evaluation strategy are presented in Section 4.1 of Massimo et al.~\cite{rizzoli2025civet}.

The results are summarized in Table~\ref{tab:understanding}. Overall, all models achieve high accuracy on both class and attribute recognition, with \interns and \llavaos attaining perfect accuracy. Notably, no model falls below $91\%$ accuracy, confirming that all are capable of recognizing the class and the attributes of our elementary objects, making them suitable candidates for the subsequent experiments in our study.

\begin{table}[ht]
    \caption{Accuracy (\%) of each model on scene understanding}
    \centering
    \begin{threeparttable}
        \begin{tabular}{lrrrr}
            \toprule
            \textbf{Model} & \textbf{Class} & \textbf{Attributes}  \\
            \midrule
            \textit{Random Baseline}
            &	25	&	17	\\
            \textit{\clip}  
            &	95	&	91	\\
            \textit{\coca}	            
            &	92	&	\textbf{100}	\\
            \textit{\intern}
            &	\textbf{100}	&	\textbf{100}	\\
            \textit{\llavai}
            &	98	&	91	\\
            \textit{\llavao}
            &	\textbf{100}	&	\textbf{100}	\\
            \textit{\paligemma}
            &	\textbf{100}	&	99	\\
            \textit{\qwen}
            &	99	&	99	\\
            \bottomrule
        \end{tabular}
        {\footnotesize \textit{Results are averaged across all 24 target objects. The best results are highlighted in \textbf{bold}} \par}
    \end{threeparttable}
    \label{tab:understanding}
\end{table}

\subsection*{Data}
We specify additional details about the data used in our experiment. The prompt template employed for counting is: \textit{“Answer with as few words as possible. How many $<$targets$>$ are there? Choose from [$<$options$>$].”} Here, \textit{$<$targets$>$} corresponds to the class and attributes of the target object, such as ``large magenta circles'' or ``small red stars''.

As the image and object size may affect the accuracy of these models, we adopt an image resolution of $672 \times 672$ for all experiments, based on previous findings~\cite{rizzoli2025civet}. For models that do not natively support this resolution, we apply model-specific pre-processing strategies (e.g., resizing followed by center cropping for \clip).

\subsection*{Evaluation Details}
All experiments for our systematic evaluation and layer-wise analysis were conducted on a single NVIDIA A100 GPU with 40 GiB of memory, using a batch size of 4. Across all settings, answers to the closed-ended questions were produced using greedy decoding.

\subsection*{Training Details}
We provide additional details of our fine-tuning experiments for full reproducibility. For each model, we first identify the optimal learning rate via hyperparameter search, with the results summarized in Table~\ref{tab:lr}. We then fine-tune the output layer for 50 epochs using AdamW, while detaching it from the embedding layer and freezing all other parameters. To accelerate training, we first precompute the hidden representations of the final layer\footnote{All models fit on a single NVIDIA A100 GPU with 40 GiB} (as the rest of the network remains fixed). By doing so, each fine-tuning experiment requires no more than 10GiB, with a batch size of 32, and can fit on a single GeForce RTX 2080 Ti GPU with 11GiB. Without our pre-processing strategy, the same experiments would require up to 95× longer training time and an NVIDIA A100 GPU with 80GiB of memory. 

\begin{table}[ht]
    \caption{Learning rate used during fine-tuning for Synthetic and Real-World data}
    \centering
    \begin{tabular}{lrrrr}
        \toprule
        \textbf{Model} & \textbf{Synthetic} & \textbf{Real-World}  \\
        \midrule
        \textit{\intern}
        &	1e-3	&	1e-4	\\
        \textit{\llavai}
        &	1e-3	&	1e-5	\\
        \textit{\llavao}
        &	1e-2	&	1e-5	\\
        \textit{\paligemma}
        &	1e-2	&	1e-4	\\
        \textit{\qwen}
        &	1e-3	&	1e-5	\\
        \bottomrule
    \end{tabular}
    \label{tab:lr}
\end{table}

\subsection*{Additional Results}
To further contextualize the magnitude of the errors produced by the evaluated VLMs, we report additional results using Mean Absolute Error (MAE) and Root Mean Squared Error (RMSE) in Table~\ref{tab:acc_mae_rmse}. A comparison between \cocas and \clips shows that, although \cocas achieves higher accuracy than \clip, it exhibits a larger RMSE. This indicates that \cocas tends to produce larger deviations when incorrect. Similarly, the lower MAE of \llavais compared to \cocas suggests that \llavai's predictions are, on average, closer to the ground truth.

\begin{table}[ht]
    \caption{Accuracy ($\%$), Mean Absolute Error (MAE), and Root Mean Squared Error (RMSE) under the baseline setting}
    \label{tab:acc_mae_rmse}
    \centering
    \begin{threeparttable}
        \begin{tabular}{lccc}
            \toprule
            \textbf{Model} & \textbf{Accuracy ↑} & \textbf{MAE ↓} & \textbf{RMSE ↓} \\
            \midrule
            \textbf{Dual-Encoders}          & & & \\
            \quad \textit{\clip}           & $20.52$ &  $2.114$  & $2.993$ \\
            \vspace{0.5em}
            \quad \textit{\coca}           & $48.16$ &  $2.057$  & $3.510$ \\
            \textbf{Encoder-Decoders}          & & & \\
            \quad \textit{\intern}         & $52.32$ &  $0.628$ & $0.990$ \\
            \quad \textit{\llavai}         & $45.66$ &  $1.012$  & $1.622$ \\
            \quad \textit{\llavao}         & $67.46$ &  $0.373$  & $0.688$ \\
            \quad \textit{\paligemma}      & $74.86$ &  $0.272$  & $0.568$ \\
            \quad \textit{\qwen}           & $64.72$ &  $0.391$  & $0.686$ \\
            \bottomrule
        \end{tabular}
        {\footnotesize \textit{Accuracy, MAE, and RMSE are computed across all 24 elementary target objects under the baseline setting. Lower MAE and RMSE indicate better counting performance.} \par}
    \end{threeparttable}
\end{table}

In addition to the error analysis, we present results illustrating the impact of varying the class and attributes of the target object in Table~\ref{tab:class_performance} and Table~\ref{tab:attribute_performance}, respectively. Consistent with the observations reported in the main manuscript, changes in the object's class have the strongest influence on counting accuracy. This effect is particularly evident for \intern, where the accuracy decreases from $61.32\%$ when counting circles to $37.84\%$ when counting squares.

\begin{table}[ht]
    \caption{Accuracy ($\%$) of each model when varying the target object's class}
    \label{tab:class_performance}
    \centering
    \begin{threeparttable}
        \begin{tabular}{lcccc}
            \toprule
            \textbf{Model} & \textbf{Squares} & \textbf{Circles} & \textbf{Triangles} & \textbf{Stars} \\
            \midrule
            \textit{\clip}      & 25.67 & 22.84 &  6.72 & 26.86 \\
            \textit{\coca}      & 37.84 & 56.40 & 50.80 & 47.60 \\
            \textit{\intern}    & 35.44 & 56.72 & 61.32 & 55.83 \\
            \textit{\llavai}    & 44.40 & 52.26 & 40.81 & 45.15 \\
            \textit{\llavao}    & 60.63 & 74.10 & 71.95 & 63.17 \\
            \textit{\paligemma} & 73.78 & 73.39 & 76.66 & 75.63 \\
            \textit{\qwen}      & 54.64 & 67.76 & 66.48 & 69.98 \\
            \bottomrule
        \end{tabular}
        {\footnotesize \textit{For each class, accuracy is computed by marginalizing across all attributes (i.e., colors).} \par}
    \end{threeparttable}
\end{table}

\begin{table}[ht]
    \caption{Accuracy ($\%$) of each model when varying the target object's attributes}
    \label{tab:attribute_performance}
    \centering
    \begin{threeparttable}
        \begin{tabular}{lcccccc}
            \toprule
            \textbf{Model} & \textbf{R} & \textbf{G} & \textbf{B} & \textbf{C} & \textbf{M} & \textbf{Y} \\
            \midrule
            \textit{\clip}      & 19.62 & 17.80 & 23.56 & 22.22 & 18.90 & 21.06 \\
            \textit{\coca}      & 51.95 & 42.94 & 46.50 & 50.10 & 49.76 & 47.70 \\
            \textit{\intern}    & 55.93 & 49.83 & 50.41 & 51.78 & 53.60 & 52.40 \\
            \textit{\llavai}    & 46.98 & 46.57 & 45.92 & 44.14 & 45.61 & 44.72 \\
            \textit{\llavao}    & 68.00 & 67.80 & 69.34 & 66.43 & 68.55 & 64.64 \\
            \textit{\paligemma} & 73.59 & 73.73 & 76.65 & 73.46 & 76.41 & 75.34 \\
            \textit{\qwen}      & 65.40 & 63.99 & 66.46 & 62.14 & 66.22 & 64.09 \\
            \bottomrule
        \end{tabular}
        {\footnotesize \textit{For each attribute, accuracy is computed by marginalizing across all classes (i.e., shapes). Notation: \textbf{R}: red, \textbf{G}: green, \textbf{B}: blue, \textbf{C}: cyan, \textbf{M}: magenta, \textbf{Y}: yellow.} \par}
    \end{threeparttable}
\end{table}

\subsection*{Layer-Wise Analysis}
\begin{figure}[ht]
    \centering
    \includegraphics[width=0.8\linewidth]{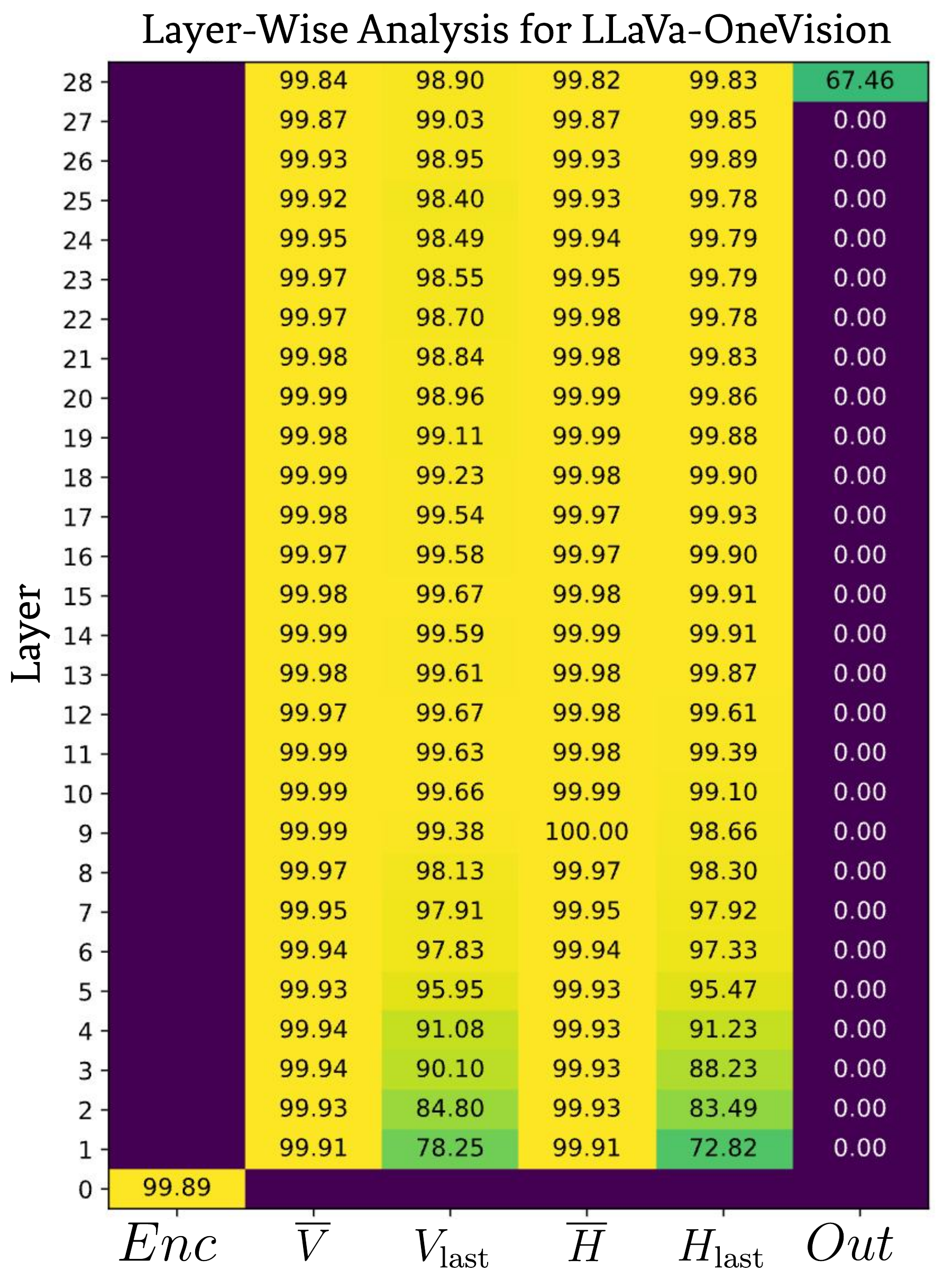}
    \caption{Layer-wise counting performance of \llavao.}
    \label{fig:layer-wise-onevision}
\end{figure}

\begin{figure}[ht]
    \centering
    \includegraphics[width=0.8\linewidth]{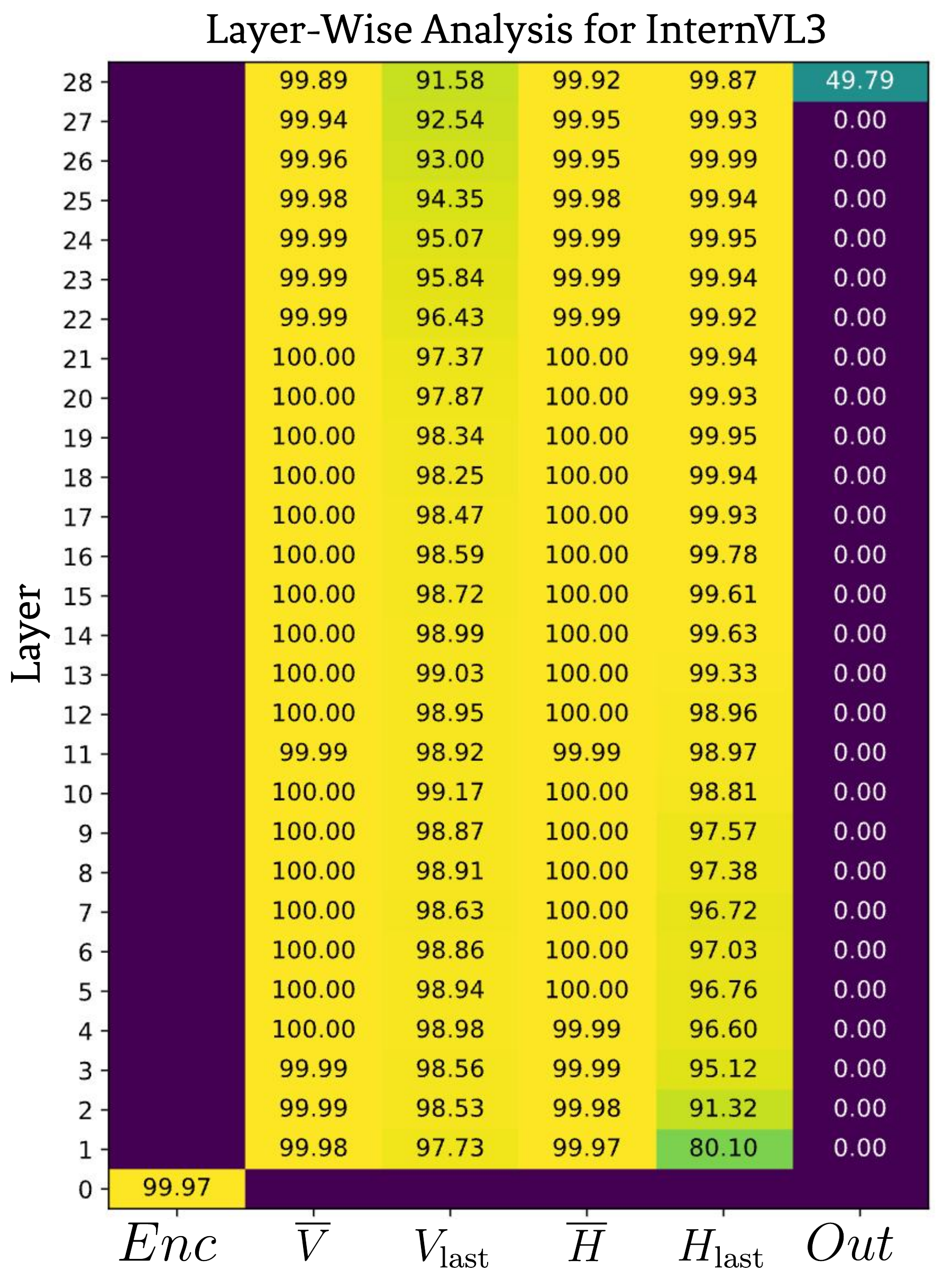}
    \caption{Layer-wise counting performance of \intern.}
    \label{fig:layer-wise-intern}
\end{figure}

\begin{figure}[ht]
    \centering
    \includegraphics[width=0.8\linewidth]{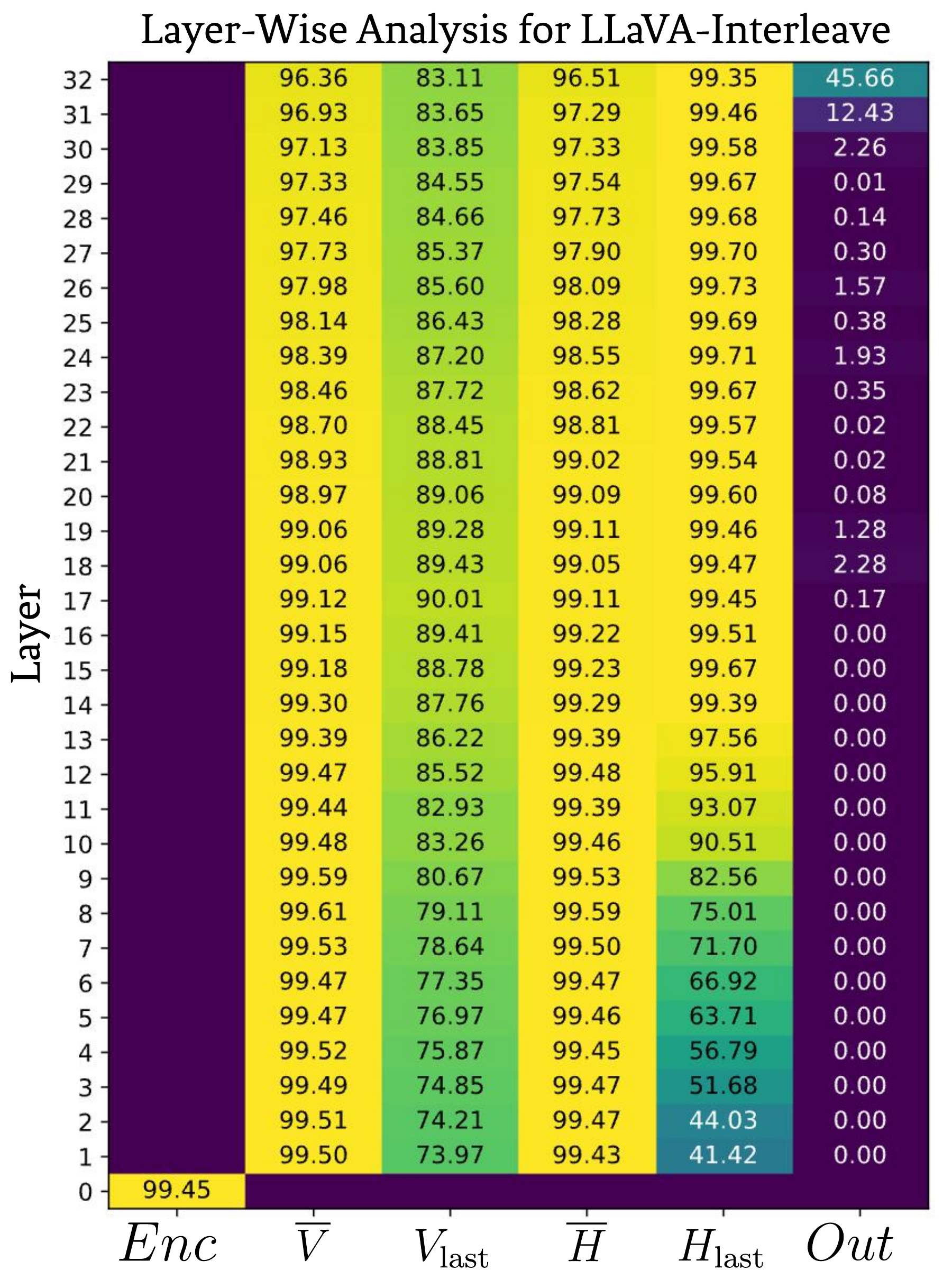}
    \caption{Layer-wise counting performance of \llavai.}
    \label{fig:layer-wise-interleave}
\end{figure}

\begin{figure}[ht]
    \centering
    \includegraphics[width=0.8\linewidth]{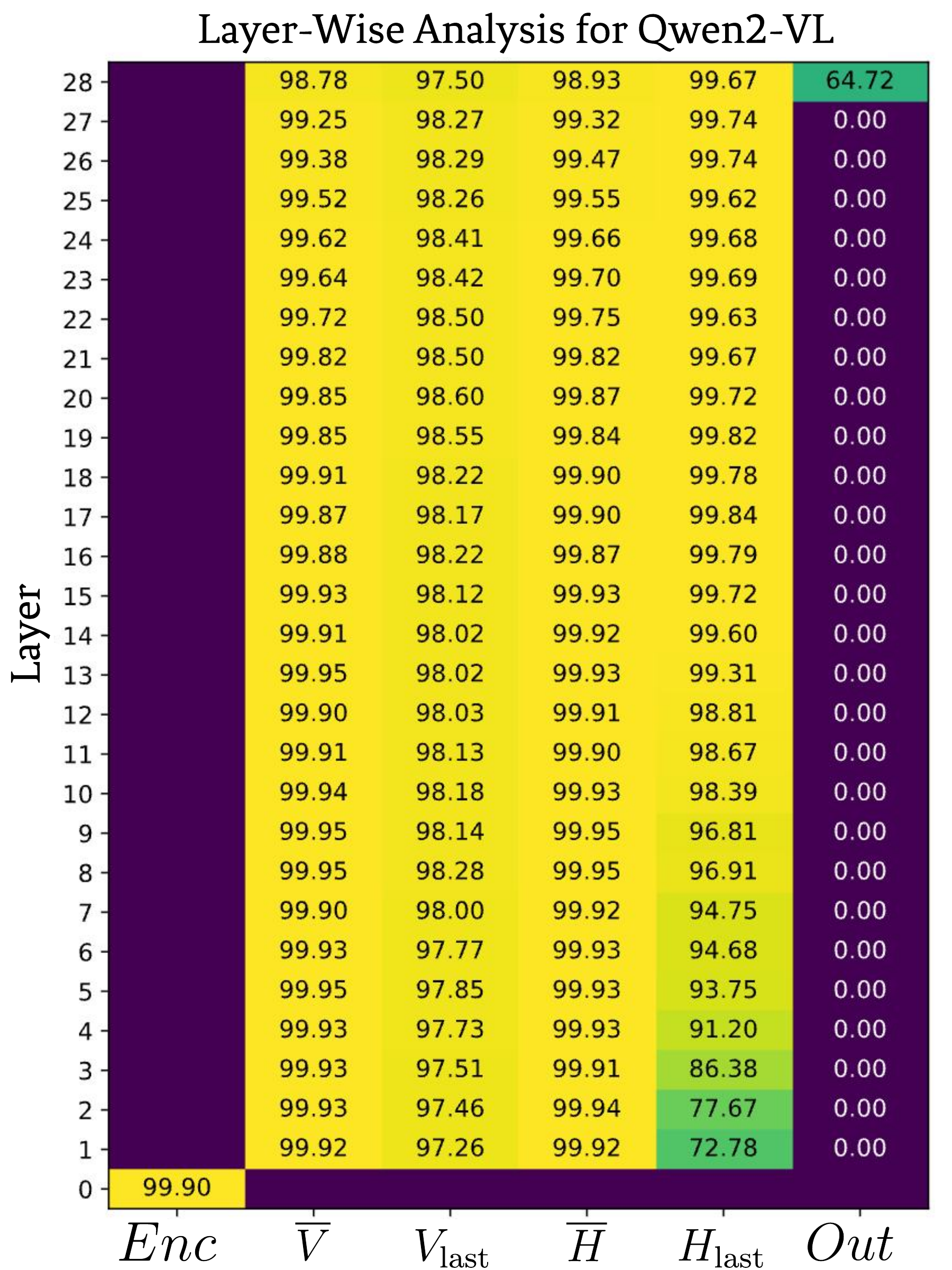}
    \caption{Layer-wise counting performance of \qwen.}
    \label{fig:layer-wise-qwen}
\end{figure}

\begin{figure}[ht]
    \centering
    \includegraphics[width=0.8\linewidth]{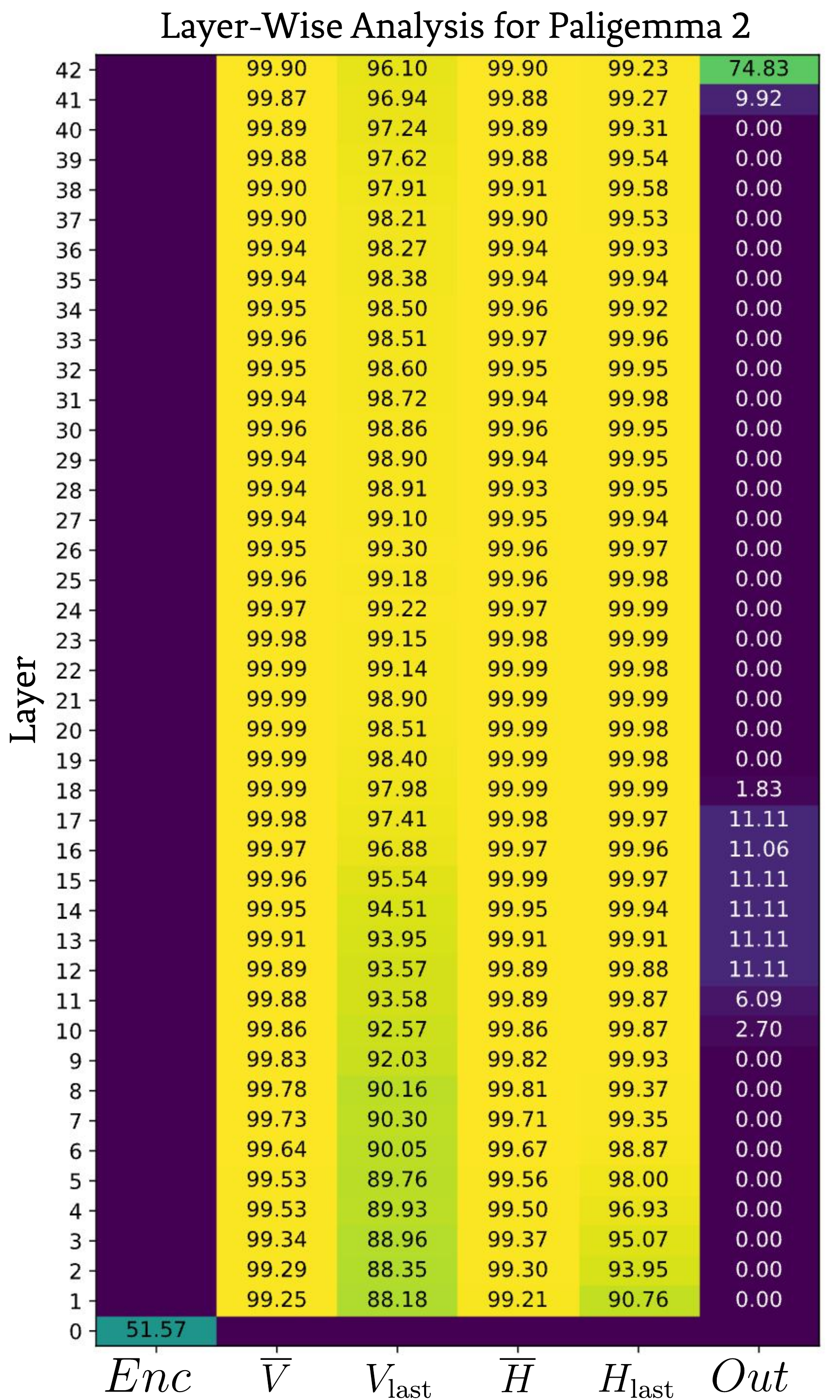}
    \caption{Layer-wise counting performance of \paligemma.}
    \label{fig:layer-wise-paligemma}
\end{figure}

Figures~\ref{fig:layer-wise-onevision}, \ref{fig:layer-wise-intern}, \ref{fig:layer-wise-interleave}, \ref{fig:layer-wise-qwen}, and \ref{fig:layer-wise-paligemma} present additional results from our layer-wise analysis for \llavao, \intern, \llavai, \qwen, and \paligemma, respectively. While most findings are consistent with those reported in the main paper, \paligemmas exhibits an interesting behavior. Specifically, Figure~\ref{fig:layer-wise-paligemma} shows that training our simple probe $P_i$ directly on $Enc$ yields suboptimal performance, achieving only $51\%$ accuracy. However, when the same representation is passed through the decoder, accuracy increases up to $99\%$. This suggests that the encoder representation is not directly linearly separable, yet the decoder appears to have learned a transformation that renders the information linearly separable.

\end{document}